\documentclass[10pt,journal,compsoc]{IEEEtran}
\ifCLASSOPTIONcompsoc
  \usepackage[nocompress]{cite}
\else
  \usepackage{cite}
\fi

\ifCLASSINFOpdf
\else
\fi

\usepackage{graphicx}
\usepackage{amsmath,amssymb} 
\usepackage{color}

\usepackage{times}
\usepackage{epsfig}
\usepackage{color}
\usepackage{multirow}
\usepackage{enumerate}
\usepackage{array}
\usepackage{paralist}
\usepackage{cite}

\newcommand\eg{\emph{e.g.}} 
\newcommand\ie{\emph{i.e.}} 
\newcommand\cf{\emph{c.f.}} 
\newcommand\etal{\emph{et al.}}

\usepackage[breaklinks=true,bookmarks=false]{hyperref}

\begin{document}
%
\title{Deeply Supervised Depth Map Super-Resolution as Novel View Synthesis}

\author{Xibin~Song,
        Yuchao~Dai,
        and~Xueying~Qin
\IEEEcompsocitemizethanks{
\IEEEcompsocthanksitem X. Song is with the School of Computer Science and Technology, Shandong University, China, Baidu Research, Beijing, China, and National Engineering Laboratory of Deep Learning Technology and Application, China. E-mail: song.sducg@gmail.com \protect\\
\IEEEcompsocthanksitem Y. Dai is  with Shaanxi Key Lab of Information Acquisition and Processing, School of Electronics and Information, Northwestern Polytechnical University, E-mail: daiyuchao@gmail.com, daiyuchao@nwpu.edu.cn. Y. Dai is the corresponding author. \protect\\
\IEEEcompsocthanksitem X. Qin is with the School of Software, Shandong University, China, E-mail: qxy@sdu.edu.cn.}
\thanks{Copyright \textcopyright~2018 IEEE. Personal use of this material is permitted. However, permission to use this material for any other purposes must be obtained from the IEEE by sending an email to pubs-permissions@ieee.org.}}


\IEEEtitleabstractindextext{
\begin{abstract}Deep convolutional neural network (DCNN) has been successfully applied to depth map super-resolution and outperforms existing methods by a wide margin. However, there still exist two major issues with these DCNN based depth map super-resolution methods that hinder the performance: i) The low-resolution depth maps either need to be up-sampled before feeding into the network or substantial deconvolution has to be used; and ii) The supervision (high-resolution depth maps) is only applied at the end of the network, thus it is difficult to handle large up-sampling factors, such as $\times 8, \times 16$. In this paper, we propose a new framework to tackle the above problems. First, we propose to represent the task of depth map super-resolution as a series of novel view synthesis sub-tasks. The novel view synthesis sub-task aims at generating (synthesizing) a depth map from different camera pose, which could be learned in parallel. Second, to handle large up-sampling factors, we present a deeply supervised network structure to enforce strong supervision in each stage of the network. Third, a multi-scale fusion strategy is proposed to effectively exploit the feature maps at different scales and handle the blocking effect. In this way, our proposed framework could deal with challenging depth map super-resolution efficiently under large up-sampling factors (\eg $\times 8, \times 16$). Our method only uses the low-resolution depth map as input, and the support of color image is not needed, which greatly reduces the restriction of our method. Extensive experiments on various benchmarking datasets demonstrate the superiority of our method over current state-of-the-art depth map super-resolution methods.
\end{abstract}

}

\maketitle

\IEEEdisplaynontitleabstractindextext

\IEEEpeerreviewmaketitle

\section{Introduction}\label{sec:introduction}
\begin{figure}[t]
  \centering
  \includegraphics[width=\linewidth]{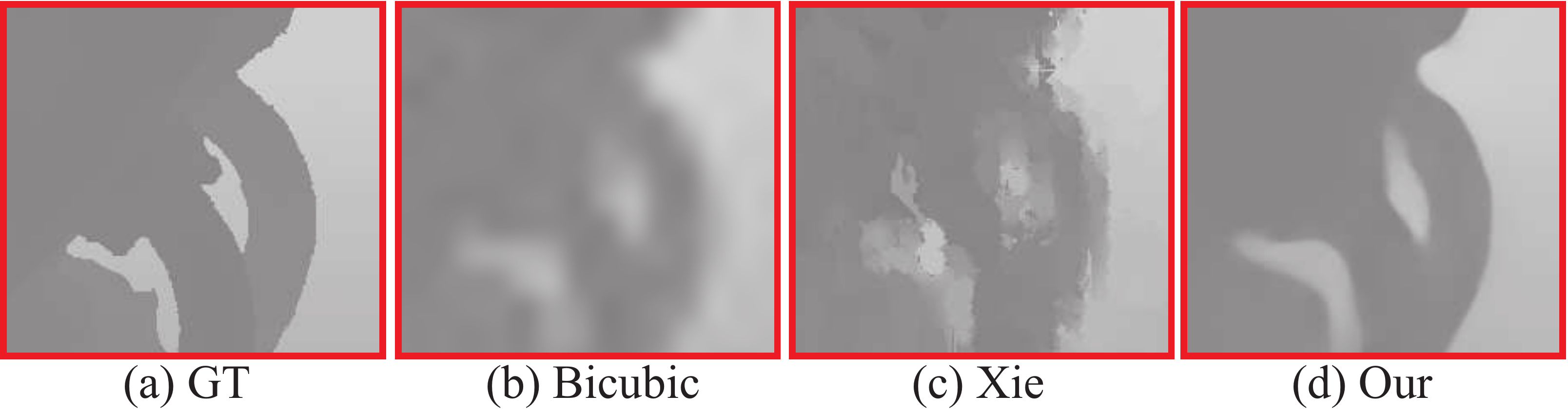} \\
  \caption{Qualitative comparison between our method and state-of-the-art methods for DSR with noisy input under an up-sampling factor $\times 16$. (a) Ground truth, (b) Bicubic, (c) Xie \etal~\cite{edge-guided-single-depth-image-super-resolution:TIP-2016} and (d) Our result.}
  \label{fig:Rank}
\end{figure}

\IEEEPARstart{D}{epth} map super-resolution (DSR) (\cf, Fig.~\ref{fig:Rank}) aims at super-resolving a high-resolution depth map from a low-resolution depth map input \cite{zhang:A-unified-scheme-for-super-resolution-depth-estimation-from-asymmetric-stereoscopic-video:tcsvt2016,song:Learning-depth-super-resolution-using-deep-convolutional-neural-network:ACCV2016,zhou:nonlocal-pixel-selection-for-multisurface-fitting-based-SR:tcsvt2014,Deng:single-image-SR-via-an-iterative-reproducing-kernel-hilbert-space-method:tcsvt2016}, which is a challenging task especially under large up-sampling factors ($\times 4$, $\times 8$, $\times 16$ and beyond). This is mainly due to the great information loss in down-sampling. For example, under the up-sampling factor of $\times 16$, 256 ($16\times 16$) depth values have to be inferred from a single depth value on average. To tackle this highly under-constrained problem, various methods have been proposed by exploiting the availability of large-scale training datasets. Even though deep convolutional neural network (DCNN) based methods have achieved great success in various vision tasks such as image deblurring~\cite{sUN:learning-a-convolutional-neural-network-for-non-uniform-motion-blur-removal:CVPR2015}\cite{Xu:Deep-convolutional-neural-network-for-image-deconvolution:NIPS2014}, image denoising~\cite{Jain:Natural-image-denoising-with-convolutional-networks:NIPS2009}\cite{Xie:Image-denoising-and-inpainting-with-deep-neural-networks:NIPS2012}, monocular depth estimation \cite{Libo-CVPR-2015,Libo-PatternRecognition-2018}, saliency prediction \cite{Survey-saliency,Jingzhang-CVPR-2018}, and even color image super-resolution (CSR)~\cite{Learning-a-deep-convolutional-network-for-image-super-resolution:ECCV-2014}\cite{Dong:Accelerating-the-super-resolution-convolutional-neural-network:ECCV2016}\cite{Kim:Accurate-image-super-resolution-using-very-deep-convolutional-networks:CVPR2016}\cite{Kim:Deeply-Recursive-Convolutional-Network-for-Image-Super-Resolution:CVPR2016}, it is only very recently that the success of DCNN in color image super-resolution~\cite{Learning-a-deep-convolutional-network-for-image-super-resolution:ECCV-2014}\cite{Kim:Accurate-image-super-resolution-using-very-deep-convolutional-networks:CVPR2016}\cite{Kim:Deeply-Recursive-Convolutional-Network-for-Image-Super-Resolution:CVPR2016} has been extended to the task of depth map super-resolution~\cite{song:Learning-depth-super-resolution-using-deep-convolutional-neural-network:ACCV2016}\cite{hui:Depth-map-super-resolution-by-deep-multi-scale-guidance:ECCV2016}\cite{riegler:A-deep-primal-dual-netwok-for-guided-depth-super-resolution:BMVC2016}\cite{Riegler:ATGV-Net:ECCV2016}. This is mainly due to the intrinsic differences between color images and depth maps, where the depth maps generally contain less textures and more sharp boundaries, and are usually degraded by noise due to the imprecise consumer depth cameras. The difficulty in capturing high-resolution depth map further increases the challenge.

Under the pipelines of current DCNN based DSR methods~\cite{song:Learning-depth-super-resolution-using-deep-convolutional-neural-network:ACCV2016}\cite{riegler:A-deep-primal-dual-netwok-for-guided-depth-super-resolution:BMVC2016}\cite{Riegler:ATGV-Net:ECCV2016}, a depth map usually needs to be up-sampled before feeding into the network. However, the up-sampled depth maps do not necessarily provide a proper good initialization for the network learning. And the problem becomes even worse when the resolution of the input low-resolution depth map is too low and the up-sampling factor is too large. Hence, to improve the representative ability of DCNN, Riegler \etal~\cite{riegler:A-deep-primal-dual-netwok-for-guided-depth-super-resolution:BMVC2016}\cite{Riegler:ATGV-Net:ECCV2016} resorted to increasing the depth of the network. Unfortunately, this deep structure may suffer from the vanishing gradient issue as the supervision is only enforced at the very end of the network. Besides, deconvolution strategy~\cite{hui:Depth-map-super-resolution-by-deep-multi-scale-guidance:ECCV2016} has also been used in DSR to improve the quality of the resultant feature maps, which can be viewed as an inverse operation of convolution. The deconvolution operator generally needs a large number of parameters. In this paper, we would like to argue that neither the hand-designed up-sampling nor the deconvolution is necessary for depth map super-resolution.

\begin{figure}
  \centering
  \includegraphics[width=\linewidth]{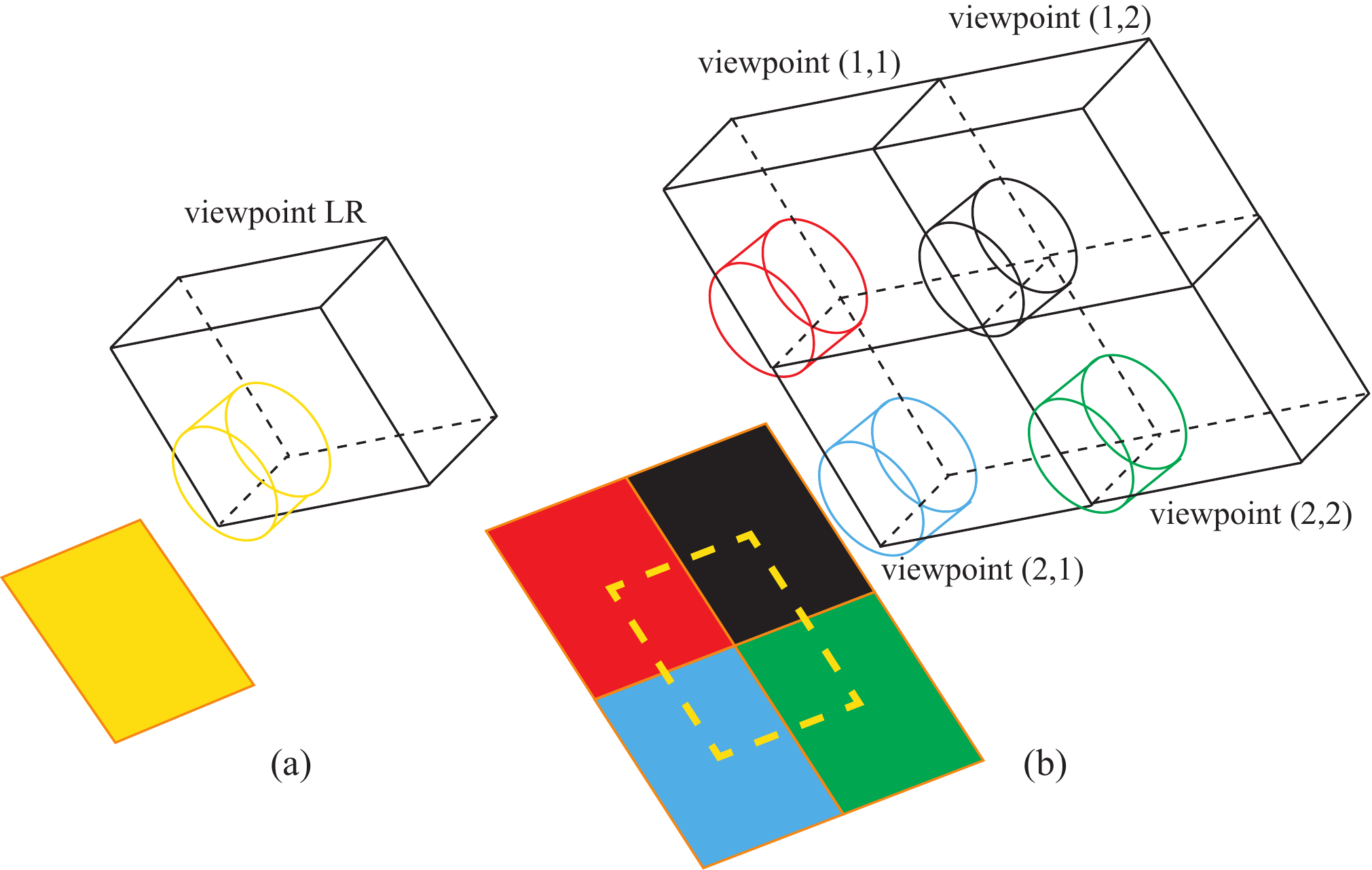} \\
  \caption{Depth map super-resolution as novel view synthesis. We illustrate the novel view synthesis process for an up-sampling factor of $\times2$. (a) shows the input single pixel which can be regarded as the input low-resolution depth map captured at $viewpoint$ $LR$; (b) shows the output four pixels which can be regarded as pixels captured from four slightly different viewpoints, respectively. The red, black, blue and green pixels are corresponding to the positions of $(1,1), (1,2), (2,1)$ and $(2,2)$. The yellow imaginary line in (b) corresponds to the yellow pixel in (a).}
  \label{virtual_camera}
\end{figure}

In this paper, we adopt a different way by representing the task of depth map super-resolution as a series of novel view synthesis sub-tasks at different positions, where each sub-task generates a depth map at a different camera pose. Take the depth map super-resolution task with an up-sampling factor $\times 2$ as an example, where we would like to generate $2\times 2=4$ new depth values from one depth measurement in the low-resolution depth map. We partition the desired high-resolution depth map with size ($H\times W$) into four parts: $\mathbf{D}_{1,1}^{HR} = \{\mathbf{D}_{2i-1,2j-1}^{HR}\}$, $\mathbf{D}_{1,2}^{HR}=\{\mathbf{D}_{2i-1,2j}^{HR}\}$, $\mathbf{D}_{2,1}^{HR} = \{\mathbf{D}_{2i,2j-1}^{HR}\}$ and $\mathbf{D}_{2,2}^{HR} = \{\mathbf{D}_{2i,2j}^{HR}\}$, where $i=1,\cdots,H/2, j=1,\cdots,W/2$. These four depth maps own the same spatial resolution and could be viewed as depth maps captured by virtual depth cameras at four different positions, which has the same resolution as the low-resolution depth map $\mathbf{D}^{LR}$ with resolution ($H/2 \times W/2$). Therefore, instead of learning a direct nonlinear mapping from $\mathbf{D}^{LR}$ to $\mathbf{D}^{HR}$, we propose to learn four separate nonlinear mappings (\eg novel view synthesis) from $\mathbf{D}^{LR}$ to $\mathbf{D}_{1,1}^{HR}$, $\mathbf{D}_{1,2}^{HR}$, $\mathbf{D}_{2,1}^{HR}$ and $\mathbf{D}_{2,2}^{HR}$ separately, where each of the nonlinear mapping could be learned through DCNN in parallel. Then the super-resolution task can be formulated as predicting the depth maps corresponding to the four different virtual cameras from the input low-resolution depth map. In this way, the input and output of each novel view synthesis task have the same resolution, which makes the network structure easy to design and implement. In Fig.~\ref{virtual_camera}, we illustrate our idea of representing depth map super-resolution as novel view synthesis.

Furthermore, to handle large up-sampling factors, a deeply supervised learning strategy is proposed. In training each sub-task, the supervision from the output could be used independently. Thus, we achieve a deeply supervised learning framework to DSR. As the supervision has been deeply enforced at different layers of the network, strong supervision is well expected. As each novel view synthesis sub-task is learned independently, there will be blocking effect between different parts, which only happens in the final output depth map. To reduce the blocking effect and effectively exploit the feature maps at different stages, we propose a multi-scale fusion strategy (MSF) to further improve the learned depth map, where inter-media depth maps at different scales are fused. Finally, to better utilize the individual information from each depth map, we impose a global depth field statistic prior (DFS) to further optimize the obtained depth maps. Our method does not need the support of color information, which reduces the restriction of corresponding color images.

As illustrated in Fig.~\ref{fig:Rank}, our method outperforms state-of-the-art DSR methods especially under large up-sampling factors ($\times 4, \times 8, \times 16$). Our main contributions can be summarized as:
\begin{compactenum}[1)]
\item We represent the task of depth map super-resolution as a series of novel view synthesis sub-tasks, where each sub-task can be efficiently solved in an end-to-end learning manner;
\item A deeply supervised learning framework is proposed to handle large up-sampling factors in depth map super-resolution, where strong supervisions are applied at different stages;
\item A multi-scale fusion strategy (MFS) and depth field statistic (DFS) are proposed to effectively exploit the feature maps at different scales and handle the blocking effect;
\item Experiments on various benchmarking datasets demonstrate the superiority of our method over state-of-the-art depth map super-resolution methods, including the DSR methods using the color images.
\end{compactenum}



\begin{figure*}
  \centering
  \includegraphics[width=\linewidth]{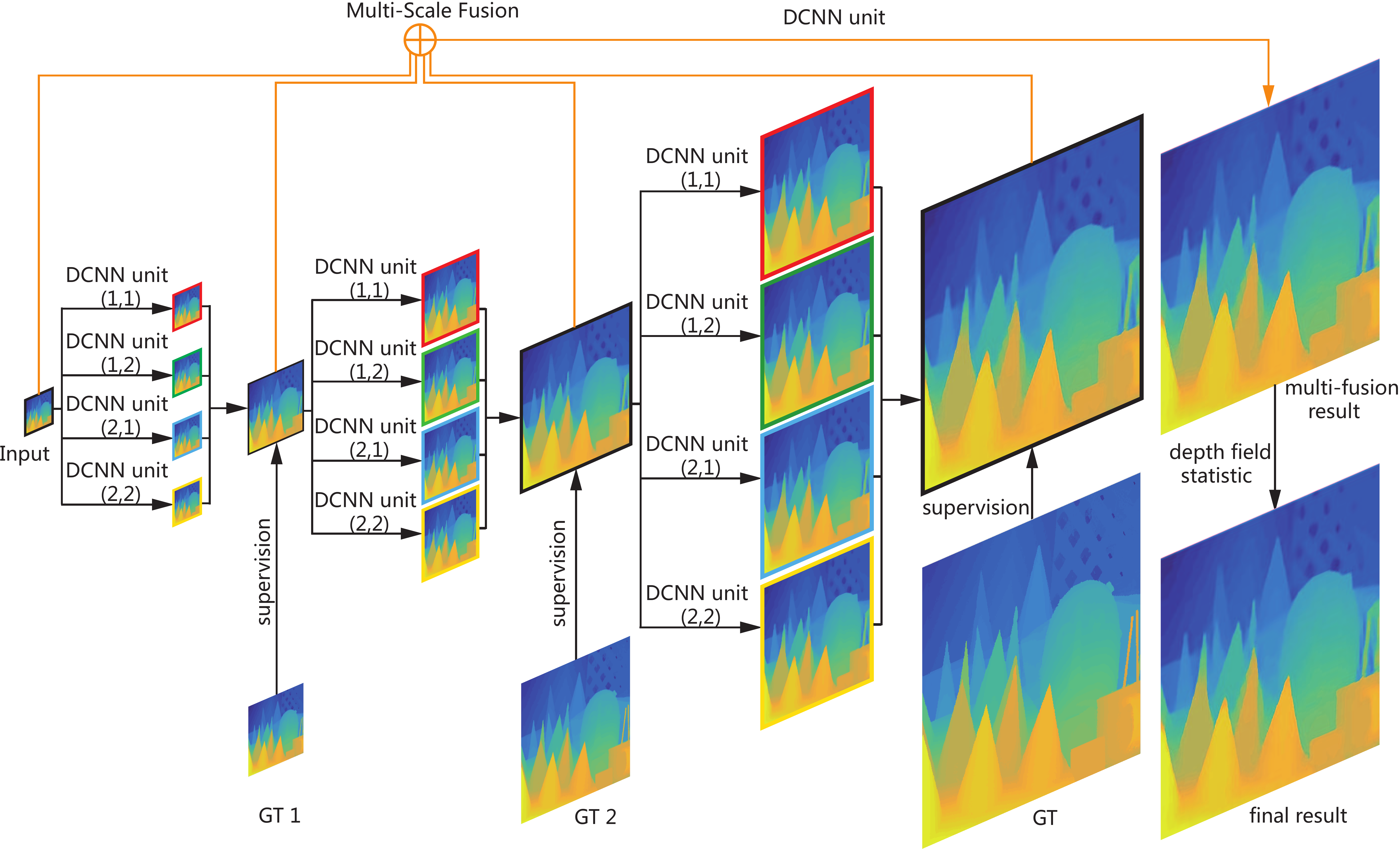} \\
  \caption{A nutshell of our depth map super-resolution method for an up-sampling factor of $\times 8$. Given an LR depth map as input, the DCNN unit is first used to train novel view synthesis sub-tasks in parallel, then a re-organization operation is utilized to obtain the up-sampled depth map. Then, the resultant depth map is regarded as input to the next stage. Deep supervision is enforced at each stage, where the supervision signals are down-sampled from the ground truth high-resolution depth maps. The results of each stage are fused by using our multi-scale fusion (MSF) strategy. Finally, a depth field statistic (DFS) prior is applied to further improve the quality of the fused depth maps.}
  \label{fig:work_flow_new}
\end{figure*}

\section{Related Work}

\subsection{Depth image super-resolution}

Depth map super-resolution (DSR) methods can be roughly classified into three categories: conventional learning based DSR, high-resolution (HR) intensity image guided DSR and deep convolutional neural network (DCNN) based DSR.

\textbf{Conventional learning based DSR:} It is proven that HR depth maps can be generated by low-resolution (LR) depth maps based on prior information. Inspired by Freeman \etal ~\cite{Freeman:Example-based-super-resolution:CGA:2002}, Aodha \etal~\cite{Patch-based-synthesis-for-single-depth-image-super-resolution:ECCV-2012} proposed a patch based MRF method to DSR by using prior information learned from depth map datasets. Horn{\'a}cek \etal~\cite{Depth-super-resolution-by-rigid-body-self-similarity-in-3d:CVPR-2013} generated HR depth maps by searching low- and high-resolution patch-pairs of arbitrary size in the depth map itself. What's more, sparse representation and dictionary learning have also been utilized in DSR. Ferstl \etal~\cite{Varitional-depth-superresolution-using-example-based-edge-representations:ICCV-2015} proposed to generate HR depth edges by learning from an external database of high and low-resolution examples, where the Total Generalized Variation (TGV) was employed as regularization. Xie \etal~\cite{edge-guided-single-depth-image-super-resolution:TIP-2016} used the MRF optimization approach to generate sharp HR depth edges from LR depth edges, where the HR depth edges were used as guidance in generating the HR depth maps. 

\textbf{HR intensity image guided DSR:} Pre-aligned HR intensity images can always provide effective guidance for DSR since they contain plenty of useful high-frequency components which assist the process of DSR. Park \etal~\cite{Park:high-quality-depth-map-upsampling-for-3D-TOF-cameras:ICCV-2011} generated HR depth maps by non-local means filter with an HR intensity image as auxiliary information. Yang \etal~\cite{Yang:Depth-Recovery-Using-an-Adaptive-Color-Guided-Auto-regressive-Model:ECCV2012} used an adaptive color-guided auto-regressive (AR) model to generate HR depth maps from LR depth maps. Ferstl \etal~\cite{Image-guided-depth-upsampling-using-anisotropic-total-generalized-variation:ICCV-2013} utilized an anisotropic diffusion tensor, which used HR color image as guidance. Kiechle \etal~\cite{Kiechle:joint-intensity-and-depth-co-sparse-analysis-model:ICCV-2013} made use of a bimodal co-sparse analysis model to generate an HR depth map from an LR depth map and an HR color image. Additionally, Matsuo \etal~\cite{Matsuo:depth-image-enchancement-using-local-tangent-plane-approximations:CVPR-2015} generated HR depth maps by using auxiliary information extracted from HR color images to compute local tangent planes in depth maps. What's more, Lu \etal~\cite{Lu:Sparse-depth-super-resolution:CVPR-2015} utilized the consistency between color images and depth maps to generate HR depth maps. However, notwithstanding the appealing results that such approaches could generate, the lack of high-resolution color images fully registered with the depth maps in many cases makes the color image guided approaches less general.

\textbf{DCNN based DSR:} The success of DCNN in high-level computer vision tasks has only been extended to DSR very recently. Song \etal~\cite{song:Learning-depth-super-resolution-using-deep-convolutional-neural-network:ACCV2016} proposed a progressive DCNN based end-to-end learning method to generate an HR depth map from an LR depth map, where SRCNN~\cite{Learning-a-deep-convolutional-network-for-image-super-resolution:ECCV-2014} was used as the mapping unit in the progressive process. Meanwhile, Riegler \etal~\cite{Riegler:ATGV-Net:ECCV2016} proposed to combine DCNN with total variations in a novel ATGV-Net to generate HR depth maps. The total variations were expressed by layers with fixed parameters. Besides, Riegler \etal~\cite{riegler:A-deep-primal-dual-netwok-for-guided-depth-super-resolution:BMVC2016} also proposed a novel DCNN based method by combining a DCNN with a non-local variational method. The corresponding HR color images were also utilized in the method. LR depth maps were all needed to be up-sampled before feeding them to the network~\cite{song:Learning-depth-super-resolution-using-deep-convolutional-neural-network:ACCV2016}\cite{riegler:A-deep-primal-dual-netwok-for-guided-depth-super-resolution:BMVC2016}\cite{Riegler:ATGV-Net:ECCV2016}. Very recently, Hui \etal~\cite{hui:Depth-map-super-resolution-by-deep-multi-scale-guidance:ECCV2016} proposed to use a multi-scale fusion strategy in a Multi-Scale Guided convolutional network for DSR with and without the guidance of the color images.

\subsection{DCNN based CSR}

The success of DCNN has been extended to color image super-resolution (CSR). Using DCNN, effective frameworks have been proposed to super-resolve the low-resolution color images.

Dong~\etal~\cite{Learning-a-deep-convolutional-network-for-image-super-resolution:ECCV-2014} proposed an end-to-end deep CNN framework to learn the nonlinear mapping between low- and high-resolution images. Based on~\cite{Learning-a-deep-convolutional-network-for-image-super-resolution:ECCV-2014}, Kim~\etal~\cite{Kim:Accurate-image-super-resolution-using-very-deep-convolutional-networks:CVPR2016} proposed to use a deeper network to represent the non-linear mapping and improved performance has been achieved. Meanwhile, in~\cite{Kim:Deeply-Recursive-Convolutional-Network-for-Image-Super-Resolution:CVPR2016}, a deeply-recursive convolutional network for CSR is proposed, which uses a deep recursive layer to obtain better results. Shi~\etal~\cite{Shi:Real-time-single-image-and-video-super-resolution-using-an-efficient-sub-pixel-CNN:CVPR2016} proposed a sub-pixel DCNN for CSR instead of using the deconvolution strategy, which utilized a sub-pixel layer to learn an array of up-sampling filters to upscale the LR feature maps to HR output. Ledig~\etal~\cite{Ledig:photo-realistic-single-image-super-resolution-using-a-generative-adversarial-network:cvpr2017} proposed a residual framework to infer photo-realistic natural images, where an adversarial loss and a content loss were used. Recently, Tai~\etal~\cite{Tai:image-super-resolution-via-deep-recursive-residual-network:cvpr2017} proposed a Deep Recursive Residual Network to mitigate the difficulty of training deep networks and used Recursive learning to control the model parameters. Lai~\etal~\cite{lai:deep-laplacian-pyramid-networks-for-fast-and-accurate-super-resolution:cvpr2017} proposed a Laplacian Pyramid Super-Resolution Network to progressively solve the problem. To handle high magnification ratios and create realistic textures, Sajjadi~\etal~\cite{Sajjadi:enhancement:single-image-SR-through-automated-texture-synthesis:iccv2017} proposed to use feed-forward fully CNN and perceptual loss to achieve automated texture synthesis. Tong~\etal~\cite{Tong:image-super-resolution-using-dense-skip-connections:iccv2017} introduced dense skip connections in a very deep network for CSR. Huang~\etal~\cite{Huang:3D-medical-images-SR:CVPR2017} proposed to utilize weakly-supervised joint convolutional sparse coding to solve the problem of multi-modal image super-resolution.

\section{Our Approach}
In this paper, we propose a new framework to tackle the problems in handling large up-sampling factors (\eg, $\times 8$, $\times 16$) and the need of deconvolution or pre-processing. First, we propose to represent the task of depth map super-resolution as a series of novel view synthesis sub-tasks. Each novel view synthesis sub-task aims at generating (synthesizing) a depth map from a slightly different camera pose, which could be learned in parallel. Second, to handle large up-sampling factors, we present a deeply supervised network structure to enforce strong supervision in each stage of the network. Third, to exploit the feature maps learned in different stages with different scales, we propose to use a multi-scale fusion (MFS) strategy, which fuses the inter-media feature maps at different scales. In this way, our framework could tackle the challenging task depth map super-resolution with large up-sampling factors efficiently. In Fig.~\ref{fig:work_flow_new}, we illustrate the whole pipeline of our method under an up-sampling factor of $\times 8$. Our method takes a low-resolution depth map as input and does not require the corresponding color image.

\subsection{Deconvolution in DCNN based DSR}
Deconvolution layer~\cite{Dong:Accelerating-the-super-resolution-convolutional-neural-network:ECCV2016}~\cite{hui:Depth-map-super-resolution-by-deep-multi-scale-guidance:ECCV2016}, as an inverse operation of convolution, is a novel operation to recover an HR depth map from an LR depth map. The deconvolution layer employs a set of deconvolution filters to up-sample the feature maps, and, the filter is convolved with the image by a stride of $1/k$, and the output is $k$ times of the input. Most of the current DCNN based depth super-resolution methods~\cite{song:Learning-depth-super-resolution-using-deep-convolutional-neural-network:ACCV2016}\cite{riegler:A-deep-primal-dual-netwok-for-guided-depth-super-resolution:BMVC2016}\cite{Riegler:ATGV-Net:ECCV2016} need the up-sampled LR depth images as input, the process of up-sampling LR depth maps can also be viewed as a deconvolution layer whose parameters are fixed. However, up-sampling the low-resolution depth map does not necessarily provide a proper initialization of the network.

Recently, Shi \etal~\cite{Shi:Real-time-single-image-and-video-super-resolution-using-an-efficient-sub-pixel-CNN:CVPR2016} proposed a sub-pixel DCNN for color image super-resolution without deconvolution. Feature maps are extracted in the LR space, and a subpixel layer which learns an array of up-sampling filters to upscale LR feature maps to HR output is utilized. Inspired by~\cite{Shi:Real-time-single-image-and-video-super-resolution-using-an-efficient-sub-pixel-CNN:CVPR2016}, we present a novel view to deal with DSR as deeply supervised novel view synthesis task.

\subsection{DSR as Novel View Synthesis}
To avoid up-sampling the LR depth maps before feeding them into the network, we represent DSR as a series of novel view synthesis sub-tasks. Rather than treating it as learning a nonlinear mapping between the low-resolution depth map and the high-resolution depth map, we decompose the desired output as a collection of low-resolution depth maps acquired at slightly different viewpoints from different ``virtual cameras''. Under this setup, each virtual camera owns the same spatial resolution which is the same as the input low-resolution depth map. For example, given a DSR task with an up-sampling factor $r$, we could decompose the DSR task to $r^2$ novel view synthesis tasks. This is because each virtual camera can be explained as having a translation between them, or forming a light field camera as illustrated in Fig.~\ref{virtual_camera}.

\begin{figure}[h]
  \centering
  \includegraphics[width=\linewidth]{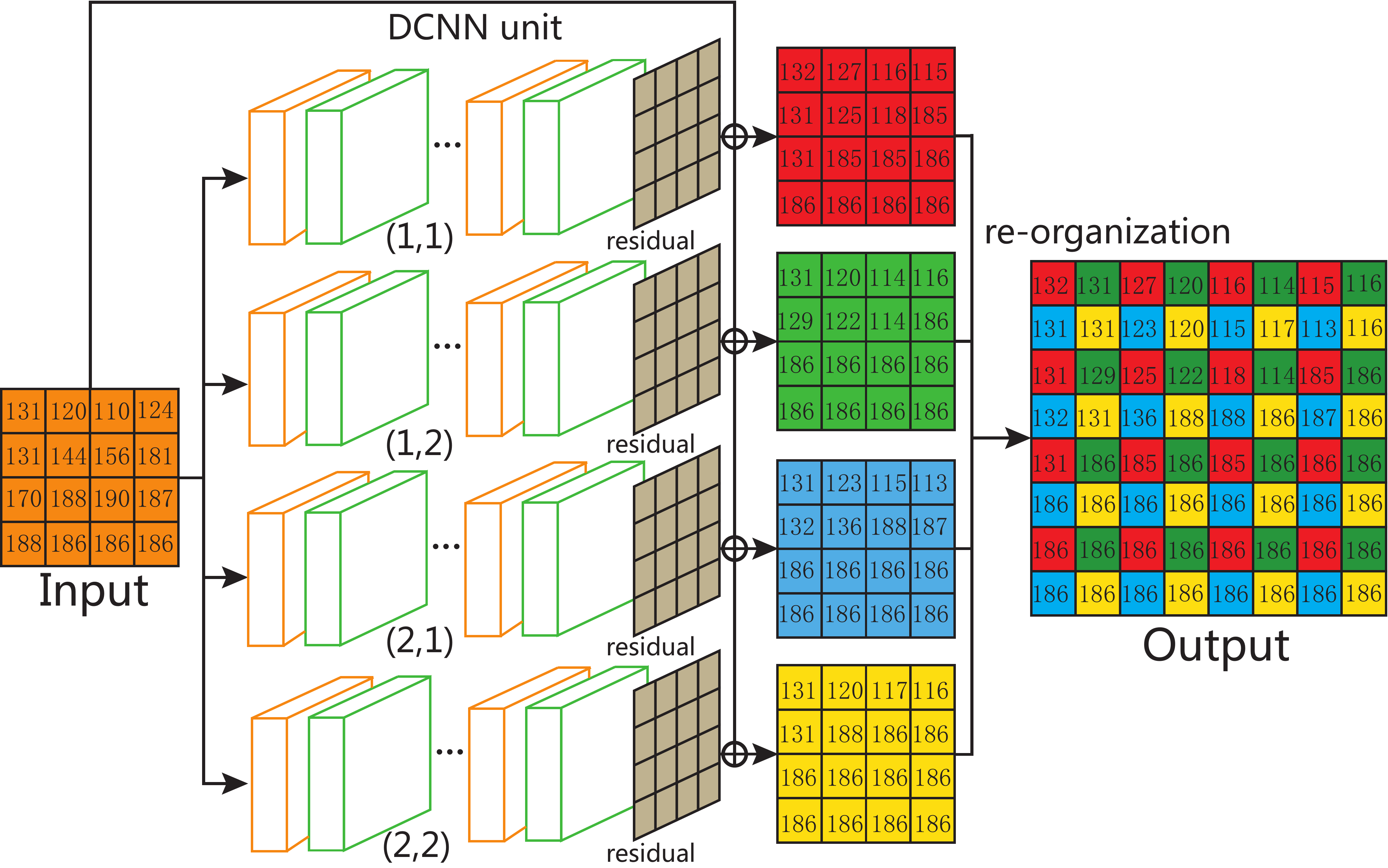} \\
  \caption{The process of novel view synthesis. The number in the grid of input denotes the depth value of each pixel. The DCNN unit is first employed to estimate the depth value of each position of novel views, then a re-organization operation is applied to generate the desired high-resolution output.}
  \label{sub_process}
\end{figure}

Specifically, as shown in Fig.~\ref{sub_process}, given the current network input, under $r=2$, we denote the task of DSR as predicting the depth maps viewed at the position of $(1,1), (1,2), (2,1)$ and $(2,2)$. Under a pure down-sampling version, the low-resolution depth map is exactly the depth map viewed at the position of $(1,1)$. Then the first novel view synthesis is directly an identical mapping. The other three tasks could be directly viewed as predicting the depth map with a motion vector, say $(1,2)$ for the horizontal direction, $(2,1)$ for the vertical direction and $(2,2)$ for the diagonal direction. With different down-sampling operators, the detailed meaning may be different, but the principle is pretty the same. The above formulation could be generalized to other up-sampling factors with ease.

Through this representation, these low-resolution depth maps of the virtual cameras could be learned in parallel. Meanwhile, as the whole DSR task has been decoupled to several different novel view synthesis sub-tasks, where for each sub-task, the output and input have the same spatial resolution. Thus, neither deconvolution nor the up-sampled input is needed in our approach.

As illustrated in Fig.~\ref{sub_process}, given an LR depth map $\mathbf{D}^{LR}$ as input, $l$ layers DCNN unit is first employed to estimate the depth values at each novel view synthesis in parallel ($\mathbf{D}_{1,1}^{OP}$, $\mathbf{D}_{1,2}^{OP}$, $\mathbf{D}_{2,1}^{OP}$ and $\mathbf{D}_{2,2}^{OP}$), where $\mathbf{D}_{i,j}^{OP}$ is the result of each sub-net. Then a re-organization operation is utilized to generate the output $\mathbf{D}^{OP}$. Note that we use a residual DCNN unit here.

Specifically, for each novel view synthesis sub-task, the process can be described as follows:
\begin{equation}
F^{i,j}_{1}(\mathbf{D}^{LR};\mathbf{W}^{i,j}_{1},\mathbf{b}^{i,j}_{1}) = \psi(\mathbf{W}^{i,j}_{1}*\mathbf{D}^{LR}+\mathbf{b}^{i,j}_{1}),
\end{equation}
\begin{equation}
	\begin{aligned}
	\mathbf{D}^{OP}_{i,j} = F^{i,j}_{l}(\mathbf{D}^{LR};\mathbf{W}^{i,j}_{1:l},\mathbf{b}^{i,j}_{1:l}) + \mathbf{D}^{LR}_{i,j}= \\ \psi(\mathbf{W}^{i,j}_{l}*F^{i,j}_{l-1}(\mathbf{D}^{LR};\mathbf{W}^{i,j}_{l-1},\mathbf{b}^{i,j}_{l-1}) + \mathbf{b}^{i,j}_{l}) + \mathbf{D}^{LR}_{i,j},
	\end{aligned}
\end{equation}
where $F$ denotes the nonlinear mapping between the LR depth map and the HR depth map. $\mathbf{W}^{i,j}$, $\mathbf{b}^{i,j}$ are the learnable network weights and biases respectively, $*$ denotes convolution. $\mathbf{D}^{LR}$ is the input LR depth map, $\psi$ is a nonlinear function and $\mathbf{D}^{OP}_{i,j}$ is the estimated novel view synthesis output for position $(i,j)$.

Then, the final output $\mathbf{D}^{HR}$ can be obtained by using the re-organization operation which is described as follows:
\begin{equation}
\mathbf{D}^{HR} = \mathcal{RO}(\mathbf{D}^{OP}_{1,1},\mathbf{D}^{OP}_{1,2},\mathbf{D}^{OP}_{2,1},\mathbf{D}^{OP}_{2,2}),
\end{equation}
where $\mathcal{RO}$ is a periodic operation that re-arranges $r^2$ low-resolution depth maps of dimension $H\times W$ to a high-resolution depth map of dimension $rH \times rW$.

\textbf{Training each sub-task:} Take an up-sampling factor of $\times 2$ as an example, the ground truth high-resolution depth map is decomposed into four sub-ground truth depth maps, (the reverse process of the re-organization operation as illustrated in Fig.~\ref{sub_process}), and in each sub-task, the LR depth map has the same resolution with the corresponding high-resolution supervision depth sub-map, which makes the DCNN easy to design and implement. Thus, neither deconvolution or up-sampling of the LR input are needed.

\subsection{Deeply Supervised Learning}
During training, besides the high-resolution target/ground truth depth map used to supervise the final output, its down-sampled versions have also been used at different stages of the depth map learning framework, denoted as $\mathbf{D}^{HR}$, $\mathbf{D}^{1HR}$, $\cdots$,$\mathbf{D}^{NHR}$, which are expressed as follows:
\begin{equation}
\begin{aligned}
\mathbf{D}^{1HR} = \downarrow_{\rho} \mathbf{D}^{HR}, \cdots, \\
\mathbf{D}^{NHR} = \downarrow_{\rho} \mathbf{D}^{(N-1)HR},
\end{aligned}
\end{equation}
where $N$ is the number of supervision stages and $\rho$ is the down-sampling factor between two consecutive stages. Here the bicubic interpolation is used as the down-sampling strategy, which is commonly used in depth map super-resolution and color image super-resolution. In this way, the aim of each sub-task can be viewed as learning the residual between two consecutive stages (two consecutive down-sampled versions of high-resolution target/ground truth) using the novel view synthesis strategy. The residual of each stage is much smaller than the residual between the input low-resolution depth map and the ground truth depth map, which does not need very deep layers to present. Hence, deeply supervised learning can effectively handle the gradient vanishing issue and obtain better results. In this way, the network could receive direct supervision in learning each factor, which could better regularize the learned depth map.

\begin{figure*}
  \centering
  \includegraphics[width=\linewidth]{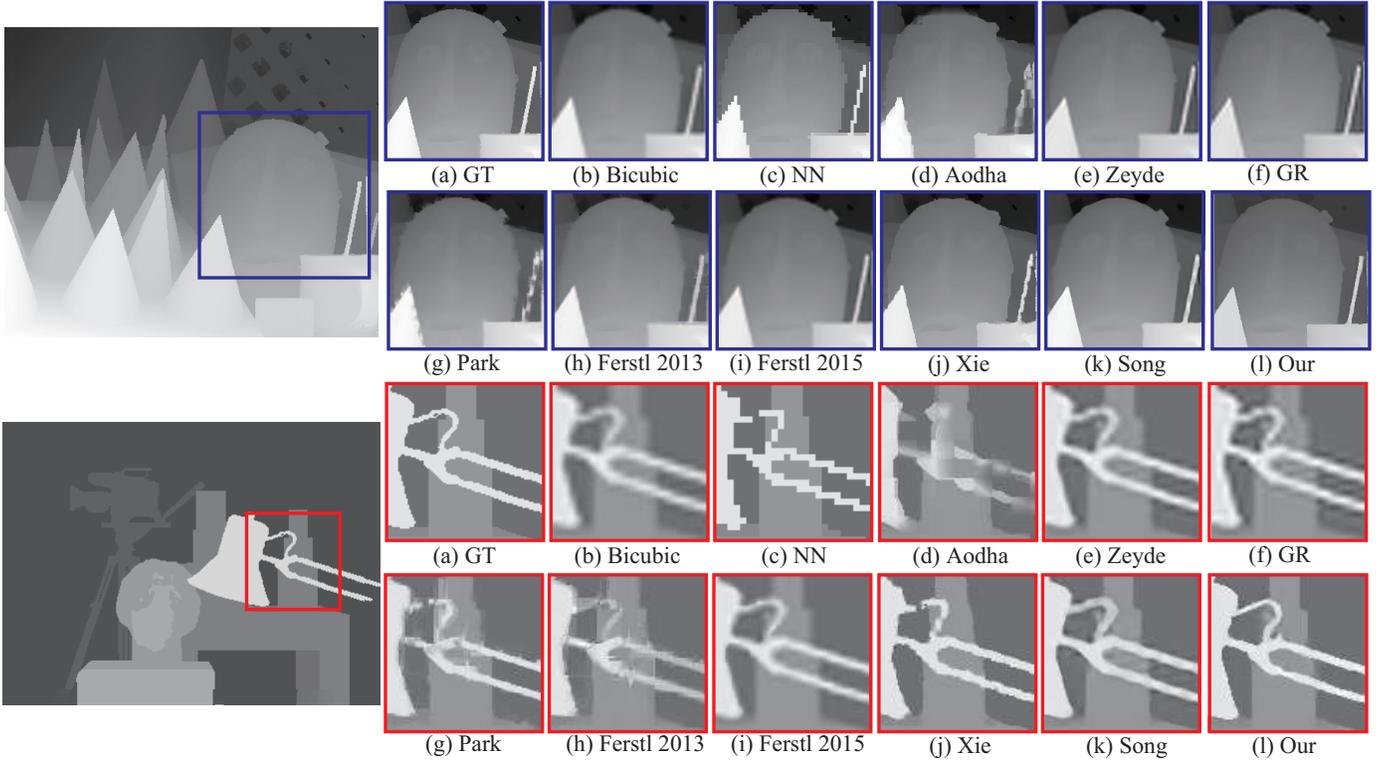} \\
  \caption{Experimental results on the Middlebury dataset (up-sampling factor $\times 4$). (a) Ground Truth. (b) Bicubic. (c) Nearest neighbor. (d) Aodha \etal~\cite{Patch-based-synthesis-for-single-depth-image-super-resolution:ECCV-2012}. (e) Zeyde \etal~\cite{On-single-image-scale-up-using-sparse-representations:CS-2012}. (f) GR~\cite{Anchored-neighborhood-regression-for-fast-example-based-super-resolution:ICCV-2013}. (g) Park \etal~\cite{Park:high-quality-depth-map-upsampling-for-3D-TOF-cameras:ICCV-2011}. (h) Ferstl \etal~\cite{Image-guided-depth-upsampling-using-anisotropic-total-generalized-variation:ICCV-2013}. (i) Ferstl \etal~\cite{Varitional-depth-superresolution-using-example-based-edge-representations:ICCV-2015}. (j) Xie \etal~\cite{edge-guided-single-depth-image-super-resolution:TIP-2016}. (k) Song \etal~\cite{song:Learning-depth-super-resolution-using-deep-convolutional-neural-network:ACCV2016}. (l) Our approach. \textbf{Best viewed on Screen.}}
  \label{result_middlebury}
\end{figure*}

\subsection{Multi-scale Fusion}
In the above deeply supervised learning structure, the ground truth depth map has been down-sampled to different resolutions to provide supervisions at different scales. While providing deep supervision and guiding the generation of high-resolution depth maps, the quality of the down-sampled depth maps has been gradually decreased due to the downsampling effect, \ie, the more downsampling has applied, the more smooth the obtained depth maps are. Therefore, the supervision depth maps actually encode the depth map details at different scales. To handle this side-effect and better utilize the inter-media feature maps, we propose a multi-scale fusion strategy (MSF) (\cf ~Fig.~\ref{fig:work_flow_new}), which fuses the feature maps (depth maps) at different stages. Specifically, the predicted depth maps are upsampled to the same resolution as the output and then are concatenated together as the input to another DCNN unit, which not only achieves multi-scale fusion of the feature maps at different scales but also effectively handles the blocking effect introduced by the individual novel view synthesis sub-tasks. By integrating both deeply supervised learning and multi-scale fusion strategies, our network is able to effectively exploit the supervision provided by the ground truth high-resolution depth map and the multi-scale feature maps.

\subsection{Network Architecture}
For each novel view synthesis sub-task, we utilize the network VDSR-Net~\cite{Kim:Accurate-image-super-resolution-using-very-deep-convolutional-networks:CVPR2016} as the DCNN unit due to its high performance in color image super-resolution. Furthermore, we propose to learn the residual between the input and the ground truth rather than learning the depth map itself. Note that the DCNN unit can be replaced by any other DCNN networks. Fig.~\ref{sub_process} shows the process of one sub-task. Taking low-resolution depth maps as input, depth maps corresponding to the positions $(1,1)$, $(1,2)$, $(2,1)$ and $(2,2)$ are trained by their corresponding DCNN units in parallel. Then, the output is obtained by re-organizing the four depth maps.

Taking the DSR task with an up-sampling factor $\times 8$ as an example, we demonstrate the network structure in Fig.~\ref{fig:work_flow_new}. The low-resolution depth maps are fed into the first stage, and the resultant depth maps are regarded as input for the next stage. Down-sampled ground truth is used as supervision to refine the depth maps at each stage. Then, the multi-scale fusion is employed to improve the quality of the resultant high-resolution depth maps, and finally, the depth field statistic (DFS) is used to further refine the resultant high-resolution depth maps.

Our method can also be extended to handle flexible up-sampling factors such as $\times 3$, $\times 5$, which can be recognized as generating depth maps of virtual cameras from 9 or 25 positions from a low-resolution depth map.

\subsection{Depth Field Statistics}
In the above end-to-end learning framework for DSR, the learned depth maps could be biased by different depth map statistics and we cannot expect that the network could learn the high-frequency information or edge information.

The distribution of a natural depth map $\mathbf{D}$ can often be modeled as a generalized Laplace distribution \cite{song:Learning-depth-super-resolution-using-deep-convolutional-neural-network:ACCV2016}, where the distribution of gradient magnitude of depth images can be well approximated with Laplacian distribution. Therefore, we propose to minimize the total variation of the depth map, \ie $\| \mathbf{D} \|_{\mathrm{TV}} \rightarrow \min$, where the total variation could be expressed in matrix form:
\begin{equation}
\|\mathbf{D} \|_{\mathrm{TV}} = \|\mathbf{P} {\mathrm{vec}} (\mathbf{D})\|_1,
\end{equation}
where $\mathbf{P}$ consists of $-1,1$ as its elements that works as the gradient operator, $\mathrm{vec}(\cdot)$ denotes the vectorization operator that transforming a matrix to a vector.

\subsection{An Energy Minimization Formulation}
To further refine the depth super-resolution results, we integrate the depth super-resolution cue from the deeply supervised DCNN and the depth field statistics (DFS), and reach the following energy minimization formulation:
\begin{equation}
\min_{\mathbf{D}} \frac{1}{2}\|\mathbf{D} - \overline{\mathbf{D}}\|_F^2 + \lambda \|\mathbf{P} {\mathrm{vec}} (\mathbf{D})\|_1,
\end{equation}
where $\overline{\mathbf{D}}$ is the depth super-resolution result from the deep neural network (novel view synthesis strategy + MFS), and $\mathbf{D}$ is the final depth map we want to generate. $\lambda$ is set as $0.7$ empirically in our experiments.
This is a convex optimization problem where a global optimal solution exists. We propose to use the iterative reweighted least squares (IRLS) \cite{chartrand:Iteratively-reweighted-algorithms-for-compressive-sensing:2008}~\cite{Hartley:Iteratively-Reweighted-Graph-Cut-for-Multi-label-MRFs-with-No-convex-Priors:CVRP:2015} to efficiently solve the problem. Given the depth map estimation result in the $it$-th iteration, the optimization for the $it+1$-th iteration can be expressed as:
\begin{equation}
\begin{aligned}
\min_{\mathbf{D}} \frac{1}{2}\|\mathbf{D} - \overline{\mathbf{D}}\|_F^2 + \lambda \sum_{i}\|\mathbf{P}_i \mathrm{vec}(\mathbf{D})\| \\
= \min_{\mathbf{D}} \frac{1}{2}\|\mathbf{D} - \overline{\mathbf{D}}\|_F^2 + \lambda \sum_{i}\frac{\|\mathbf{P}_i \mathrm{vec}(\mathbf{D})\|^2}{\|\mathbf{P}_i \mathrm{vec}(\mathbf{D}^{(it)})\|},
\end{aligned}
\end{equation}
which could be equivalently expressed as:
\begin{equation}
\begin{aligned}
\min_{\mathbf{D}} \frac{1}{2}\|\mathbf{D} - \overline{\mathbf{D}}\|_F^2 + \lambda \sum_{i}\| \frac{\mathbf{P}_i}{\sqrt{\| \mathbf{P}_i \mathrm{vec}(\mathbf{D}^{(it)})\|}} \mathrm{vec}(\mathbf{D})\|^2.
\end{aligned}
\end{equation}

Denote $\mathbf{E}_i^{(it)} = \frac{\mathbf{P}_i}{\sqrt{\| \mathbf{P}_i \mathrm{vec}(\mathbf{D}^{(it)})\|}}$, \ie, the row-wise reweighted version of $\mathbf{P}_i$ and $\mathbf{E}^{(it)} = \left[\mathbf{E}_i^{(it)}\right]$, we have:
\begin{equation}
\begin{aligned}
\mathbf{D}^{(it+1)} = \arg\min_{\mathbf{D}} \frac{1}{2}\|\mathbf{D} - \overline{\mathbf{D}}\|_F^2 + \lambda \| \mathbf{E}^{(it)} \mathrm{vec}(\mathbf{D}) \|_F^2.
\end{aligned}
\end{equation}

The above least squares problem owns a closed-form solution.

{\footnotesize
\begin{table*}[h]
\scriptsize
\begin{center}

\caption{Quantitative evaluation under clean depth map input. The $RMSE$ is calculated for different SOTA methods for clean Middlebury dataset for up-sampling factors of $\times{2}$, $\times{4}$ and $\times{8}$. $Our$ denotes our deeply supervised novel view synthesis strategy, $MSF$ means the multi-scale fusion strategy, $DFS$ means the depth field statistic prior. The best result is highlighted and the second best is underlined.}
\label{tab:mb_rmse_2_4_8}

\begin{tabular}{ |c|c|c|c|c|c|c|c|c|c|c|c|c| }
\hline
\multirow{2}{4em}{Method}  &  \multicolumn{4}{| c |}{$\times 2$} & \multicolumn{4}{| c |}{$\times 4$} & \multicolumn{4}{| c |}{$\times 8$} \\ \cline{2-13}
 & Cones & Teddy & Tsukuba & Venus & Cones & Teddy & Tsukuba & Venus & Cones & Teddy & Tsukuba & Venus \\  \hline
NN  & 4.4622 & 3.2363 & 9.2305 & 2.1298 & 6.0054 & 4.5466 & 12.9083 & 2.9333 & 7.5937 & 6.2416 & 18.4786 & 4.4645\\ 
Bicubic   & 2.5245 & 1.9495 & 5.7828 & 1.3119 & 3.8635 & 2.8930 & 8.7103 & 1.9403 & 5.3000 & 4.2423 & 13.3220 & 2.8948 \\ \hline

Park et al.~\cite{Park:high-quality-depth-map-upsampling-for-3D-TOF-cameras:ICCV-2011}   & 2.8497 & 2.1850 & 6.8869 & 1.2584 & 6.5447 & 4.3366 & 12.1231 & 2.2595 & 8.0078 & 6.3264 & 17.6225 & 3.4086 \\
Yang et al.~\cite{Yang:Depth-Recovery-Using-an-Adaptive-Color-Guided-Auto-regressive-Model:ECCV2012} & 2.4214 & 1.8941 & 5.6312 & 1.2368 & 5.1390 & 4.0660 & 13.1748 & 2.7559 & 5.1390 & 4.0660 & 13.1748 & 2.7559\\

Ferstl et al.~\cite{Image-guided-depth-upsampling-using-anisotropic-total-generalized-variation:ICCV-2013}  & 3.1651 & 2.4208 & 6.9988 & 1.4194 & 3.9968 & 2.8080 & 10.0352 & 1.6643 & N/A & N/A & N/A & N/A\\
JID~\cite{Kiechle:joint-intensity-and-depth-co-sparse-analysis-model:ICCV-2013}  & 1.7451 & 1.2681 & 3.7415 & 0.6879 & 3.0369 & 1.8043 & 5.9028 & 0.9625 & \underline{4.5929} & 2.9342 & 10.0800 & 1.2684 \\

Yang et al.~\cite{Image-super-resolution-via-sparse-representation:TIP-2010}   & 2.8384 & 2.0079 & 6.1157 & 1.3777 & 3.9546 & 3.0908& 8.2713 & 1.9850 & 5.3176 & 4.0447 & 13.0340 & 2.8140 \\ 
Zeyde et al.~\cite{On-single-image-scale-up-using-sparse-representations:CS-2012} & 1.9539 & 1.5013 & 4.5276 & 0.9305 & 3.2232 & 2.3527 & 7.3003 & 1.4751 & 4.8945 & 3.5670 & 11.9758 & 2.2879\\ 
GR~\cite{Anchored-neighborhood-regression-for-fast-example-based-super-resolution:ICCV-2013} & 2.3742 & 1.8010 & 5.4059 & 1.2153 & 3.5728 & 2.7044 & 8.0645& 1.8175 & 5.0603 & 3.8137 & 12.3357 & 2.6384\\ 
ANR~\cite{Anchored-neighborhood-regression-for-fast-example-based-super-resolution:ICCV-2013} & 2.1237 & 1.6054 & 4.8169 & 1.0566 & 3.3156 & 2.4861 & 7.4895 & 1.6449 & 4.9904 & 3.6666 & 12.1035 & 2.4653\\ 
NE+LS & 2.0437 & 1.5256 & 4.6372 & 0.9697 & 3.2868 & 2.4210 & 7.3404 & 1.5225 & 5.0948 & 3.6195 & 12.1448 & 2.3967\\ 
NE+NNLS & 2.1158 & 1.5771 & 4.7287 & 1.0046 & 3.4362 & 2.4887 & 7.5344 & 1.6291 & 4.9906 & 3.6957 & 12.2283 & 2.4647\\ 
NE+LLE & 2.1437 & 1.6173 & 4.8719 & 1.0827 & 3.3414 & 2.4905 & 7.5528 & 1.6449 & 4.9572 & 3.6916 & 12.1652 & 2.5202\\ 
Aodha et al.~\cite{Patch-based-synthesis-for-single-depth-image-super-resolution:ECCV-2012} & 4.3185 & 3.2828 & 9.1089 & 2.2098 & 12.6938 & 4.1113 & 12.6938 & 2.6497 & N/A & N/A & N/A & N/A\\ 
Horn{\'a}cek et al.~\cite{Depth-super-resolution-by-rigid-body-self-similarity-in-3d:CVPR-2013} & 3.7512 & 3.1395 & 8.8070 & 2.0383 & 5.4898 & 5.0212 & 11.1101 & 3.5833 & N/A & N/A & N/A & N/A\\ 
Huang et al.~\cite{Huang:Single-image-super-resolution-from-transformed-self-exemplars:CVPR2015}   & 4.6273 & 3.4293 & 10.0766 & 2.1653 & 6.2723 & 4.8346 & 13.7645 & 3.0606 & 6.1629 & 6.6235 & 10.6618 & 4.1399\\
Schultery et al.~\cite{schulter:Fast-and-accurate-image-upscaling-with-super-resolution-forests:CVPR2015} & 1.9199 & 1.5545 & 4.2400 & 1.0185 & 2.9859 & 2.3793 & 6.8026 & 1.5477 & N/A & N/A & N/A & N/A\\
Ferstl et al.~\cite{Varitional-depth-superresolution-using-example-based-edge-representations:ICCV-2015} & 2.2139 & 1.7205 & 5.3252 & 1.1230 & 3.5680 & 2.6474 & 7.5356 & 1.7771 & N/A & N/A & N/A & N/A\\ 
Xie et al.~\cite{edge-guided-single-depth-image-super-resolution:TIP-2016} & 2.7338 & 2.4911 & 6.3534 & 1.6390 & 4.4087 & 3.2768 & 9.7765 & 2.3714 & N/A & N/A & N/A & N/A \\
Wang et al.~\cite{Wang:Deep-networks-for-image-super-resolution-with-sparse-prior:ICCV2015} & 1.8895 & 1.4074 & 3.8789 & 0.8935 & 2.9263 & 2.0638 & 6.0356 & 1.2640 & 4.8933 & 3.0437 & 9.8942 & 1.8618 \\ \hline

MSLaplas~\cite{lai:deep-laplacian-pyramid-networks-for-fast-and-accurate-super-resolution:cvpr2017} & 1.5119 & 1.1992 & 3.3239 & 0.7370 & 3.0733 & 1.8076 & 4.9121 & 0.9380 & 5.2976 & 2.9100 & \textbf{9.4433} & 1.4769 \\

Laplas~\cite{lai:deep-laplacian-pyramid-networks-for-fast-and-accurate-super-resolution:cvpr2017} & 1.8810 & 1.4653 & 4.4722 & 0.8572 & 3.2320 & 2.0716 & 6.4812 & 1.1994 & 5.0544 & 2.8592 & 9.8536 & 1.2186 \\

Song et al.~\cite{song:Learning-depth-super-resolution-using-deep-convolutional-neural-network:ACCV2016} & 1.4356 & 1.1974 & 2.9841 & 0.5592 & 2.9789 & 1.8006 & 6.1422 & 0.8796 & \textbf{4.5887} & 2.8850 & 11.6231 & 1.7082 \\ 

MS-Net~\cite{hui:Depth-map-super-resolution-by-deep-multi-scale-guidance:ECCV2016} & 1.1000 & 0.8220 & 2.4720 & \underline{0.2590} & 2.7700 & 1.5330 & 4.9960 & \underline{0.4220} & 5.2170 & \underline{2.8740} & 9.9860 & \textbf{0.8810} \\

ATGV-Net~\cite{Riegler:ATGV-Net:ECCV2016} & 1.0021 & \underline{0.8155} & 2.3846 & \textbf{0.1991} & 2.9293 & \textbf{1.5029} & 6.6327 & \textbf{0.3764} & N/A & N/A & N/A & N/A \\

VDSR-Net~\cite{Kim:Accurate-image-super-resolution-using-very-deep-convolutional-networks:CVPR2016} & 0.9339 & 0.8548 & 1.6934 & 0.3934 & 2.3831 & 1.5469 & 4.5902 & 0.5616 & 4.9893 & 2.9458 & 10.8818 & 1.3126 \\ \hline

$Our$ & \underline{0.8991} & \underline{0.7745} & \underline{1.6786} & 0.2779 & 2.2233 & 1.5427 & 4.4539 & 0.5504 & 4.9932 & 2.8483 & \underline{9.8526} & 1.1017 \\

$Our$+$MSF$ & 0.9213 & 0.7891 & 1.7181 & 0.3206 & \underline{2.2080} & 1.5174 & \underline{4.4089} & 0.4978 & 4.9369 & \underline{2.8055} & 9.9407 & 1.0854 \\
$Our$+$MSF$+$DFS$ & \textbf{0.8757} & \textbf{0.7613} & \textbf{1.6505} & 0.2784 & \textbf{2.1956} & \underline{1.5173} & \textbf{4.3839} & 0.4987 & 4.9318 & \textbf{2.8046} & 9.9206 & \underline{1.0839} \\ \hline

\end{tabular}

\end{center}
\end{table*}
}

{\footnotesize
\begin{table*}[!h]
\scriptsize
\begin{center}

\caption{Quantitative evaluation under up-sampling factors $\times 2, \times 4, \times 8$. Note that the input is non-noisy. The $SSIM$ is calculated for different SOTA methods on the Middlebury dataset under up-sampling factors of $\times{2}$, $\times{4}$ and $\times{8}$. The best result is highlighted and the second best is underlined.}
\label{tab:mb_ssim_2_4_8}

\begin{tabular}{ |c|c|c|c|c|c|c|c|c|c|c|c|c| }
\hline
\multirow{2}{4em}{Method}  &  \multicolumn{4}{| c |}{$\times 2$} & \multicolumn{4}{| c |}{$\times 4$} & \multicolumn{4}{| c |}{$\times 8$} \\ \cline{2-13}
 & Cones & Teddy & Tsukuba & Venus & Cones & Teddy & Tsukuba & Venus & Cones & Teddy & Tsukuba & Venus\\  \hline
NN & 0.9645 & 0.9696 & 0.9423 & 0.9888 & 0.9360 & 0.9450 & 0.9003 & 0.9800 & 0.8996 & 0.9199 & 0.8387 & 0.9634 \\ 
Bicubic & 0.9720 & 0.9771 & 0.9536 & 0.9909 & 0.9538 & 0.9619 & 0.9205 & 0.9845 & 0.9314 & 0.9442 & 0.8564 & 0.9771 \\ \hline
Park et al.~\cite{Park:high-quality-depth-map-upsampling-for-3D-TOF-cameras:ICCV-2011}   & 0.9452 & 0.9610 & 0.9052 & 0.9811 & 0.9321 & 0.9510 & 0.8756 & 0.9799 & 0.9231 & 0.9426 & 0.8409 & 0.9792\\
Yang et al.~\cite{Yang:Depth-Recovery-Using-an-Adaptive-Color-Guided-Auto-regressive-Model:ECCV2012} & 0.9833 & 0.9850 & 0.9721 & 0.9946 & 0.9629 & 0.9697 & 0.9322 & 0.9882 & 0.9370 & 0.9488 & 0.8633 & 0.9773\\
Ferstl et al.~\cite{Image-guided-depth-upsampling-using-anisotropic-total-generalized-variation:ICCV-2013}  & 0.9755 & 0.9795 & 0.9576 & 0.9938 & 0.9625 & 0.9707 & 0.9245 & 0.9901 & N/A & N/A & N/A & N/A\\

JID.~\cite{Kiechle:joint-intensity-and-depth-co-sparse-analysis-model:ICCV-2013} & 0.9913 & 0.9922 & 0.9904 & 0.9983 & 0.9811 & 0.9833 & 0.9751 & 0.9971 & 0.9612 & 0.9691 & 0.9441 & 0.9941 \\

Yang et al.~\cite{Image-super-resolution-via-sparse-representation:TIP-2010} & 0.9473 & 0.9564 & 0.9072 & 0.9805 & 0.9482 & 0.9566 & 0.9015 & 0.9816 & 0.9339 & 0.9465 & 0.8662 & 0.9771 \\ 
Zeyde et al.~\cite{On-single-image-scale-up-using-sparse-representations:CS-2012} & 0.9655 & 0.9717 & 0.9438 & 0.9886 & 0.9604 & 0.9628 & 0.9147 & 0.9884 & 0.9385 & 0.9503 & 0.8718 & 0.9816 \\ 
GR~\cite{Anchored-neighborhood-regression-for-fast-example-based-super-resolution:ICCV-2013} & 0.9587 & 0.9656 & 0.9314 & 0.9862 & 0.9500 & 0.9592 & 0.9012 & 0.9817 & 0.9320 & 0.9454 & 0.8581 & 0.9761 \\ 
ANR~\cite{Anchored-neighborhood-regression-for-fast-example-based-super-resolution:ICCV-2013} & 0.9630 & 0.9693 & 0.9400 & 0.9879 & 0.9391 & 0.9452 & 0.8731 & 0.9806 & 0.9350 & 0.9478 & 0.8659 & 0.9784 \\ 
NE+LS & 0.9623 & 0.9692 & 0.9391 & 0.9887 & 1.6977 & 0.9514 & 0.9574 & 0.9042 & 0.9367 & 0.9493 & 0.8681 & 0.9799 \\ 
NE+NNLS & 0.9640 & 0.9707 & 0.9426 & 0.9883 & 0.9424 & 0.9499 & 0.8872 & 0.9820 & 0.9345 & 0.9472 & 0.8635 & 0.9785\\ 
NE+LLE & 0.9588 & 0.9658 & 0.9405 & 0.9837 & 0.9270 & 0.9331 & 0.8794 & 0.9641 & 0.9344 & 0.9462 & 0.8650 & 0.9759 \\ 
Aodha et al.~\cite{Patch-based-synthesis-for-single-depth-image-super-resolution:ECCV-2012} & 0.9606 & 0.9690 & 0.9364 & 0.9874 & 0.9392 & 0.9520 & 0.9080 & 0.9822 & N/A & N/A & N/A & N/A\\ 
Horn{\'a}cek et al.~\cite{Depth-super-resolution-by-rigid-body-self-similarity-in-3d:CVPR-2013} & 0.9696 & 0.9719 & 0.9461 & 0.9895 & 0.9501 & 0.9503 & 0.9137 & 0.9789 & N/A & N/A & N/A & N/A\\ 
Huang et al.~\cite{Huang:Single-image-super-resolution-from-transformed-self-exemplars:CVPR2015} & 0.9582 & 0.9673 & 0.9301 & 0.9875 & 0.9360 & 0.9425 & 0.8821 & 0.9784 & 0.9280 & 0.9254 & 0.9027 & 0.9712\\
Ferstl et al.~\cite{Varitional-depth-superresolution-using-example-based-edge-representations:ICCV-2015} & 0.9866 & 0.9884 & 0.9766 & 0.9963 & 0.9645 & 0.9716 & 0.9413 & 0.9893 & N/A & N/A & N/A & N/A\\ 
Xie et al.~\cite{edge-guided-single-depth-image-super-resolution:TIP-2016} & 0.8932 & 0.9012 & 0.9053 & 0.9300 & 0.8885 & 0.8927 & 0.8405 & 0.9175 & N/A  & N/A & N/A & N/A\\
Wang et al.~\cite{Wang:Deep-networks-for-image-super-resolution-with-sparse-prior:ICCV2015} & 0.9891 & 0.9907 & 0.9866 & 0.9966 & 0.9758 & 0.9798 & 0.9640 & 0.9937 & 0.9475 & 0.9618 & 0.9132 & 0.9878 \\ \hline

MSLaplas~\cite{lai:deep-laplacian-pyramid-networks-for-fast-and-accurate-super-resolution:cvpr2017} & 0.9940 & 0.9909 & 0.9926 & 0.9984 & 0.9822 & 0.9867 & 0.9797 & 0.9976 & 0.9521 & 0.9664 & \textbf{0.9333} & 0.9927 \\

Laplas~\cite{lai:deep-laplacian-pyramid-networks-for-fast-and-accurate-super-resolution:cvpr2017} & 0.9894 & 0.9946 & 0.9849 & 0.9975 & 0.9774 & 0.9824 & 0.9657 & 0.9963 & 0.9516 & 0.9658 & 0.9291 & 0.9950 \\

Song.et al~\cite{song:Learning-depth-super-resolution-using-deep-convolutional-neural-network:ACCV2016} & 0.9915 & 0.9918 & 0.9905 & 0.9989 & 0.9783 & 0.9831 & 0.9666 & 0.9976 & 0.9510 & 0.9679 & 0.9051 & 0.9903\\

MS-Net~\cite{hui:Depth-map-super-resolution-by-deep-multi-scale-guidance:ECCV2016} & 0.9952 & 0.9953 & 0.9930 & 0.9993 & 0.9817 & 0.9860 & 0.9746 & 0.9987 & 0.9511 & 0.9652 & 0.9312 & 0.9967 \\

VDSR-Net~\cite{Kim:Accurate-image-super-resolution-using-very-deep-convolutional-networks:CVPR2016} & 0.9947 & 0.9941 & 0.9970 & 0.9982 & 0.9823 & 0.9840 & 0.9784 & 0.9971 & 0.9481 & 0.9611 & 0.9174 & 0.9916 \\ \hline
$Our$ & \textbf{0.9958} & \textbf{0.9953} & \underline{0.9972} & \underline{0.9991} & 0.9855 & \textbf{0.9886} & 0.9820 & 0.9979 & 0.9539 & 0.9663 & 0.9316 & 0.9953 \\
$Our + MFS$ & 0.9954 & 0.9950 & 0.9969 & 0.9988 & \underline{0.9860} & \underline{0.9856} & \underline{0.9829} & \underline{0.9980} & \underline{0.9543} & \underline{0.9669} & 0.9307 & \underline{0.9953}  \\
$Our + MFS + DFS$ & \underline{0.9957} & \underline{0.9951} & \textbf{0.9972} & \textbf{0.9991} & \textbf{0.9861} & 0.9855 & \textbf{0.9832} & \textbf{0.9980} & \textbf{0.9546} & \textbf{0.9670} & \underline{0.9316} & \textbf{0.9954}  \\ \hline

\end{tabular}

\end{center}
\end{table*}
}

\subsection{Implementation Details}\label{implementation details}
\textbf{DCNN unit:} In this paper, we use the VDSR-Net~\cite{Kim:Accurate-image-super-resolution-using-very-deep-convolutional-networks:CVPR2016} with 10 convolutional layers as the DCNN unit. Each convolutional filter has the size of $3\times3$ for the novel view synthesis sub-task and $5\times 5$ for the multi-scale fusion strategy, and each hidden layer of the network has $64$ feature maps. For each DCNN unit, the learning rate varies from $0.1$ to $0.0001$, and the momentum is chosen as $0.9$. The stepwise decrease (4 steps with learning rate multiplier $\gamma = 0.1$) as the learning policy, and adjustable gradient clipping strategy~\cite{Kim:Accurate-image-super-resolution-using-very-deep-convolutional-networks:CVPR2016} is used.

{\footnotesize
 \begin{table*}[h]
 \scriptsize
 \begin{center}


  \caption{Quantitative results under noisy input (up-sampling factor $\times 4$). The $RMSE$ is calculated for different SOTA methods for the Middlebury dataset (2005 and 2014) and the Laserscan dataset using noisy input. The explanation is same as Table~\ref{tab:mb_rmse_2_4_8}.}
\begin{tabular}{ |c|c|c|c|c|c|c|c|c|c|c|c|}
\hline
\multirow{2}{4em}{Method}  &  \multicolumn{11}{| c |}{$\times 4$} \\ \cline{2-12}
 & Jadeplant & Motorcycle & Playtable & Flower & Art & Books & Laundry & Reindeer & Scan21 & Scan30 & Scan42 \\  \hline

 Bicubic & 3.0186 & 3.4961 & 2.5275 & 3.3368 & 3.9117 & 2.2887 & 2.9908 & 3.140 & 2.1792 & 2.1401 & 3.7099  \\
NN  & 3.6726 & 4.2625 & 3.0908 & 4.0940 & 4.7907 & 2.7146 & 3.5863 & 3.7993 & 3.1733 & 3.0431 & 5.8794 \\
Park et al.~\cite{Park:high-quality-depth-map-upsampling-for-3D-TOF-cameras:ICCV-2011}  & 3.7906 & 4.1823 & 2.2619 & 3.7799 & 4.1322 & 2.2075 & 3.0175 & 3.1835 & N/A & N/A & N/A\\
Zeyde et al.~\cite{On-single-image-scale-up-using-sparse-representations:CS-2012} & 3.0823 & 2.9615 & 1.8113 & 2.8241 & 3.8967 & 1.7817 & 2.9839 & 2.9304 & 2.3399 & 2.3371 & 3.3357 \\
Yang et al.~\cite{Image-super-resolution-via-sparse-representation:TIP-2010}  & 3.1787 & 3.0837 & 1.8748 & 2.9407 & 3.8967 & 1.7817 & 2.9839 & 2.9304 & 2.2071 & 2.1705 & 3.7650 \\
ANR~\cite{Anchored-neighborhood-regression-for-fast-example-based-super-resolution:ICCV-2013}  & 3.0660 & 2.9520 & 1.8175 & 2.8172 & 9.4621 & 8.8228 & 8.0385 & 8.7093 & 2.5350 &2.5171 & 3.6840 \\
GR~\cite{Anchored-neighborhood-regression-for-fast-example-based-super-resolution:ICCV-2013}  & 3.0660 & 2.9520 & 1.8175 & 2.8172  & 9.5812 & 8.8838 & 8.1310 & 8.8010 & 2.7205 & 2.6990 & 3.9682\\
Aodha et al.~\cite{Patch-based-synthesis-for-single-depth-image-super-resolution:ECCV-2012}  & N/A & N/A & N/A & N/A & 3.8967 & 1.7817 & 2.9839 & 2.9304 & 2.3156 & 2.1225 & 4.0006 \\
Ferstl et al.~\cite{Image-guided-depth-upsampling-using-anisotropic-total-generalized-variation:ICCV-2013} & 3.6394 & 3.8577 & 2.2640 & 3.4801 & 4.9346 & 4.5651 & 6.9055 & 4.6487 & N/A & N/A & N/A \\
JID~\cite{Kiechle:joint-intensity-and-depth-co-sparse-analysis-model:ICCV-2013}  & 2.5476 & 2.0540 & 1.2785 & 1.8310 & 2.8314 & 2.2602 & 2.5027 & 2.6417 & N/A & N/A & N/A \\
Huang et al.~\cite{Huang:Single-image-super-resolution-from-transformed-self-exemplars:CVPR2015}  & 3.6394 & 3.8577 & 2.2640 & 3.4801 & 3.7537 & 3.1474 & 3.4566 & 3.5258 & 2.6827 & 2.7017 & 3.2517 \\
Xie et al.~\cite{edge-guided-single-depth-image-super-resolution:TIP-2016}  & N/A & N/A & N/A & N/A & 3.7995 & 2.0742 & 2.6162 & 2.9992 & 2.3233 & 2.2411 & 4.0130 \\
Wang et al.~\cite{Wang:Deep-networks-for-image-super-resolution-with-sparse-prior:ICCV2015}  & 3.3520 & 3.3191 & 2.8827 & 3.0703 & 3.5072 & 2.7229 & 3.0895 & 3.1753 & 2.1460 & 2.1314 & 3.0442\\
AGTV-Net~\cite{Riegler:ATGV-Net:ECCV2016}  & N/A & N/A & N/A & N/A & 2.9800 & 1.7200 & N/A & N/A  & N/A & N/A & N/A \\

VDSR-Net~\cite{Kim:Accurate-image-super-resolution-using-very-deep-convolutional-networks:CVPR2016} & 1.9091 & 2.0237 & 1.3521 & 1.8153 & 2.0570 & 1.0768 & 1.5213 & 1.6711 & 1.4566 & 1.3233 & 1.8078 \\ \hline

$Our$ & 1.8282 & 1.9518 & 1.2695 & 1.7667 & 1.9441 & 0.9675 & 1.3915 & 1.5935 & 1.3201 & 1.2636 & 1.7610\\
$Our$+$MSF$ & \underline{1.7902} & \underline{1.9240} & \underline{1.2328} & \underline{1.7371} & \underline{1.9019} & \underline{0.9346} & \underline{1.3390} & \underline{1.5658} & \underline{1.3149} & \underline{1.2498} & \underline{1.6984} \\

$Our$+$MSF$+$DFS$ & \textbf{1.7771} & \textbf{1.9138} & \textbf{1.2236} & \textbf{1.7240} & \textbf{1.8896} & \textbf{0.9230} & \textbf{1.3301} & \textbf{1.5554} & \textbf{1.2866} & \textbf{1.2264} & \textbf{1.6778}  \\ \hline

 \end{tabular}

 \label{ls_mb_noisy_x4}
 \end{center}
 \end{table*}
 }

  {\footnotesize
\begin{table*}[!h]
\scriptsize
\begin{center}

\caption{Quantitative evaluation under noisy depth input. The $SSIM$ is calculated for different SOTA methods on the noisy Middlebury and the Laserscanner datasets under an up-sampling factor of $\times{4}$. The best result is highlighted and the second best is underlined.}
\label{tab:mb_ls_ssim_4_noise}

\begin{tabular}{ |c|c|c|c|c|c|c|c|c|c|c|c| }
\hline
\multirow{2}{4em}{Method}  &  \multicolumn{11}{| c |}{$\times 4$} \\ \cline{2-12}
 & Jadeplant & Motorcycle & Playtable & Flower & Art & Books & Laundry & Reindeer & Scan21 & Scan30 & Scan42\\  \hline

 NN & 0.8949 & 0.9332 & 0.9389 & 0.9424 & 0.9171 & 0.9418 & 0.9325 & 0.9342 & 0.9702 & 0.9707 & 0.9567 \\
 Bicubic & 0.9164 & 0.9595 & 0.9646 & 0.9653 & 0.9448 & 0.9634 & 0.9563 & 0.9581 & 0.9716 & 0.9722 & 0.9590 \\ \hline
 Park et al.~\cite{Park:high-quality-depth-map-upsampling-for-3D-TOF-cameras:ICCV-2011} & 0.9258 & 0.9762 & 0.9859 & 0.9824 & 0.9551 & 0.9717 & 0.9661 & 0.9679 & N/A & N/A & N/A\\
Zeyde et al.~\cite{On-single-image-scale-up-using-sparse-representations:CS-2012} & 0.9321 & 0.9799 & 0.9865 & 0.9839 & 0.9337 & 0.9408 & 0.9361 & 0.9378 & 0.9592 & 0.9592 & 0.9506\\
Yang et al.~\cite{Image-super-resolution-via-sparse-representation:TIP-2010} & 0.9311 & 0.9801 & 0.9866 & 0.9845 & 0.9486 & 0.9656 & 0.9597 & 0.9607 & 0.9725 & 0.9732 & 0.9599 \\
ANR~\cite{Anchored-neighborhood-regression-for-fast-example-based-super-resolution:ICCV-2013} & 0.9321 & 0.9791 & 0.9859 & 0.9831 & 0.9213 & 0.9300 & 0.9248 & 0.9264 & 0.9510 & 0.9513 & 0.9396 \\
GR~\cite{Anchored-neighborhood-regression-for-fast-example-based-super-resolution:ICCV-2013} & 0.9321 & 0.9791 & 0.9859 & 0.9831 & 0.9068 & 0.9197 & 0.9129 & 0.9146 & 0.9423 & 0.9429 & 0.9274 \\
Aodha et al.~\cite{Patch-based-synthesis-for-single-depth-image-super-resolution:ECCV-2012} & N/A & N/A & N/A & N/A & 0.9714 & 0.9869 & 0.9764 & 0.9828 & 0.9857 & 0.9873 & 0.9828\\
Ferstl et al.~\cite{Image-guided-depth-upsampling-using-anisotropic-total-generalized-variation:ICCV-2013} & 0.9237 & 0.9704 & 0.9796 & 0.9760 & 0.9631 & 0.9663 & 0.9365 & 0.9733 & N/A & N/A & N/A\\
JID~\cite{Kiechle:joint-intensity-and-depth-co-sparse-analysis-model:ICCV-2013} & 0.9421 & 0.9914 & 0.9932 & 0.9937 & 0.9287 & 0.9286 & 0.9244 & 0.9261 & N/A & N/A & N/A\\
Huang et al.~\cite{Huang:Single-image-super-resolution-from-transformed-self-exemplars:CVPR2015} & 0.8834 & 0.9078 & 0.9063 & 0.9170 & 0.9074 & 0.9077 & 0.9014 & 0.9047 & 0.9359 & 0.9350 & 0.9301\\
Xie et al.~\cite{edge-guided-single-depth-image-super-resolution:TIP-2016} & N/A & N/A & N/A & N/A & 0.9567 & 0.9706 & 0.9710 & 0.9685 & 0.9718 & 0.9716 & 0.9643\\
Wang et al.~\cite{Wang:Deep-networks-for-image-super-resolution-with-sparse-prior:ICCV2015} & 0.9077 & 0.9349 & 0.9333 & 0.9455 & 0.9320 & 0.9353 & 0.9288 & 0.9318 & 0.9655 & 0.9657 & 0.9582\\
VDSR-Net~\cite{Kim:Accurate-image-super-resolution-using-very-deep-convolutional-networks:CVPR2016} & 0.9885 & 0.9905 & 0.9921 & 0.9927 & 0.9868 & 0.9914 & 0.9876 & 0.9901 & 0.9920 & 0.9926 & 0.9924 \\ \hline
$Our$ & 0.9411 & 0.9917 & 0.9931 & 0.9936 & 0.9885 & 0.9927 & 0.9894 & 0.9913 & \underline{0.9935} & \underline{0.9935} & \underline{0.9936}\\
$Our + MFS$ & \underline{0.9898} & \underline{0.9917} & \underline{0.9931} & \underline{0.9936} & \underline{0.9886} & \underline{0.9927} & \underline{0.9895} & \underline{0.9913} & 0.9933 & 0.9934 & 0.9935 \\
$Our + MFS + DFS$ & \textbf{0.9902} & \textbf{0.9920} & \textbf{0.9934} & \textbf{0.9940} & \textbf{0.9889} & \textbf{0.9930} & \textbf{0.9898} & \textbf{0.9915} & \textbf{0.9937} & \textbf{0.9938} & \textbf{0.9940}\\ \hline

\end{tabular}

\end{center}
\end{table*}
}

{\footnotesize
 \begin{table*}[h]
 \scriptsize
 \begin{center}


  \caption{Quantitative results for large upscaling factor $\times 16$. The $RMSE$ is calculated for different SOTA methods for the Middlebury dataset (2005 and 2014) and Laserscan dataset using noisy input. The explanation is same as Table~\ref{tab:mb_rmse_2_4_8}.}

\begin{tabular}{ |c|c|c|c|c|c|c|c|c|c|c|c| }
\hline
\multirow{2}{4em}{Method}  &  \multicolumn{11}{| c |}{$\times 16$} \\ \cline{2-12}
 & Jadeplant & Motorcycle & Playtable & Flower & Art & Books & Laundry & Reindeer & Scan21 & Scan30 & Scan42 \\  \hline

Bicubic & 6.4744 & 7.5870 & 4.9166 & 5.9961 & 8.7821 & 4.0171 & 6.2251 & 6.2469 & 5.6547 & 6.0234 & 7.7719 \\
NN & 5.4158 & 6.3775 & 4.0157 & 7.3735 & 10.023 & 4.6584 & 7.0470 & 7.2392 & 9.1224 & 6.6863 & 6.5055 \\
Park et al.~\cite{Park:high-quality-depth-map-upsampling-for-3D-TOF-cameras:ICCV-2011}& 6.5770 & 8.2173 & 4.5044 & 6.3069 & 9.1964 & 3.7853 & 6.4360 & 6.0934 & N/A & N/A & N/A \\
Zeyde et al.~\cite{On-single-image-scale-up-using-sparse-representations:CS-2012}& 5.3660 & 5.9962 & 3.5302 & 5.5972 & 6.6357 & 3.4477 & 4.6541 & 5.0603 & 4.4270 & 4.0878 & 7.2210  \\
Yang et al.~\cite{Image-super-resolution-via-sparse-representation:TIP-2010}& 5.5559 & 6.2265 & 3.6571 &  5.8026 & 6.6079 & 2.9578 & 4.3285 & 4.8519 & 4.3007 & 3.8619 & 7.2310  \\
ANR~\cite{Anchored-neighborhood-regression-for-fast-example-based-super-resolution:ICCV-2013}& 5.3135 & 5.9480 & 3.5046 & 5.5557 & 6.7198 & 3.6173 & 4.8064 & 5.2210 & 4.5430 & 4.1952 & 7.2291 \\
GR~\cite{Anchored-neighborhood-regression-for-fast-example-based-super-resolution:ICCV-2013}& 5.3135 & 5.9480 & 3.5046 & 5.5557 & 6.5905 & 3.5932 & 4.7439 & 5.1316 & 4.4347 & 4.1142 & 7.0916 \\
JID~\cite{Kiechle:joint-intensity-and-depth-co-sparse-analysis-model:ICCV-2013} & 5.5170 & 5.5709 & 3.2141 & 5.0051 & 6.3400 & 3.0937 & 5.9453 & 5.3907 & N/A & N/A & N/A \\
Wang et al.~\cite{Wang:Deep-networks-for-image-super-resolution-with-sparse-prior:ICCV2015}& 5.2034 & 5.6034 & 3.8110 & 4.6530 & 6.2276 & 3.3479 & 4.4003 & 4.7777 & 3.4242 & 3.1981 & 5.4845\\
VDSR-Net~\cite{Kim:Accurate-image-super-resolution-using-very-deep-convolutional-networks:CVPR2016} & 5.1063 & 6.3280 & 3.4919 & 4.7088 & 6.7515 & 2.9180 & 4.6333 & 4.8409 & 3.7227 & 3.3443 & 6.1008 \\ \hline

$Our$ & 4.6949 & 5.2480 & 2.8377 & 3.6106  &  5.7697 & 2.0427 & 3.5901 & 3.9111 & 2.8423 & \underline{2.6082} & 4.4192 \\
$Our$+$MSF$ & \underline{4.6067} & \underline{5.1246} & \underline{2.7952} & \underline{3.5814} & \underline{5.7122} & \underline{2.0422} & \textbf{3.5603} & \underline{3.8838} & \underline{2.7718} & \textbf{2.5853} & \underline{4.3061} \\

$Our$+$MSF$+$DFS$ & \textbf{4.5960} & \textbf{5.1157} & \textbf{2.7914} & \textbf{3.5750} & \textbf{5.7067} & \textbf{2.0348} & \underline{3.5624} & \textbf{3.8775} & \textbf{2.7637} & 2.6088 & \textbf{4.2957}\\ \hline

 \end{tabular}

 \label{ls_mb_noisy_x16}
 \end{center}
 \end{table*}
 }

 \textbf{Training Data:} $115$ depth maps from the Middlebury stereo dataset~\cite{Evaluation-of-cost-functions-for-stereo-matching:CVPR-2007}\cite{Learning-conditional-random-fields-for-stereo:CVPR-2007}\cite{A-taxonomy-and-evaluation-of-dense-two-frame-stereo-correspondence-algorithms:IJCV-2002} (25 images), the Sintel dataset~\cite{Butler:A-Naturalistic-open-source-movie-for-optical-flow-evaluation:ECCV:2012} (60 images) and the synthetic New Tsukuba dataset~\cite{Martull:Realistic-CG-stereo-image-dataset-with-ground-truth-disparity-maps:ICPR2012} (30 images) are collected to construct our dataset, $100$ depth maps are used for training and $15$ depth maps are used for validation. Using these depth maps as ground-truth high-resolution depth maps $\mathbf{D}^{GT}$, the input low-resolution depth maps $\mathbf{D}^{LR}$ are generated by $\mathbf{D}^{LR} = \downarrow_\rho \mathbf{D}^{GT}$, where $\rho$ is the down-sampling factor and bicubic interpolation is used as the down-sampling strategy. Note that to accelerate the training time of larger up-sampling factors, such as $\times 4$, $\times 8$ and $\times 16$, parameters of $\times 2$ (model of deeply supervised novel view synthesis part) are used to initialize the parameters of the first stage of these larger up-sampling factors. For larger up-sampling factors, in each DCNN unit of the first stage, the learning rate varies from 0.01 to 0.0001, and the stepwise decrease (3 steps with learning rate multiplier $\gamma = 0.1$) as the learning policy. Using a Titan X GPU (Pascal), training for the task of depth map super-resolution under up-sampling factors $\times 2$, $\times 4$, $\times 8$ and $\times 16$ roughly takes 3 hours, 4 hours, 5 hours and 6 hours, respectively.

 \begin{figure*}
\centering
\includegraphics[width=\linewidth]{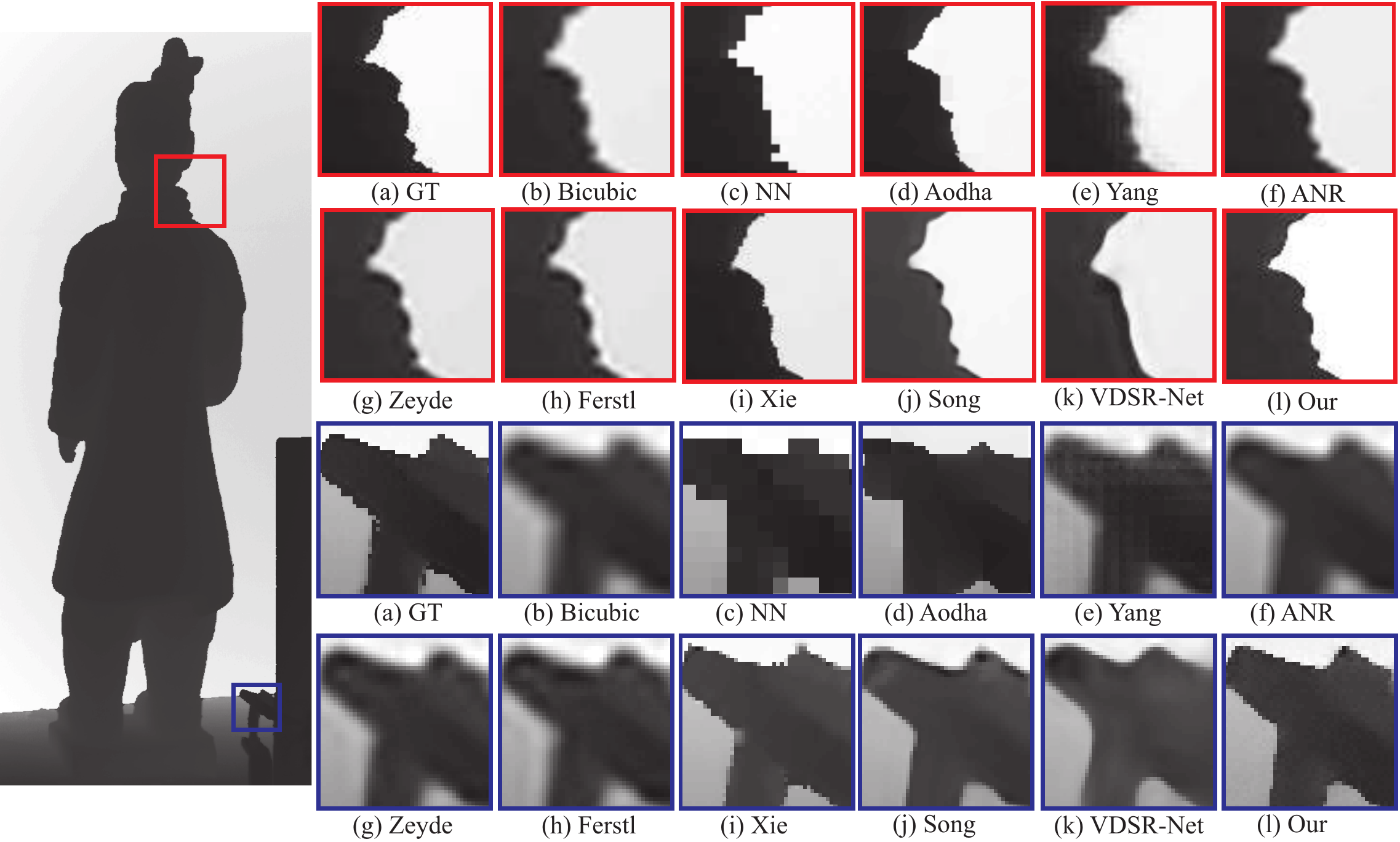} \\
  \caption{Experimental results on the clean Laserscann dataset~\cite{Patch-based-synthesis-for-single-depth-image-super-resolution:ECCV-2012} ($Scan 42$) (up-sampling factor $\times 4$). (a) Ground truth; (b) Bicubic; (c) Nearest Neighbor; (d) Aodha \etal~\cite{Patch-based-synthesis-for-single-depth-image-super-resolution:ECCV-2012}; (e) Yang \etal~\cite{Image-super-resolution-via-sparse-representation:TIP-2010}; (f) ANR~\cite{Anchored-neighborhood-regression-for-fast-example-based-super-resolution:ICCV-2013}; (g) Zeyde \etal~\cite{On-single-image-scale-up-using-sparse-representations:CS-2012}; (h) Ferstl \etal~\cite{Varitional-depth-superresolution-using-example-based-edge-representations:ICCV-2015}; (i) Xie \etal~\cite{edge-guided-single-depth-image-super-resolution:TIP-2016}; (j) Song \etal~\cite{song:Learning-depth-super-resolution-using-deep-convolutional-neural-network:ACCV2016}; (k) VDSR-Net~\cite{Kim:Accurate-image-super-resolution-using-very-deep-convolutional-networks:CVPR2016}; (l) Our results. \textbf{Best viewed on screen.}}
  \label{fig:ls_clean_x8}
\end{figure*}

\section{Experimental Results}


In this section, we present an extensive experimental evaluation of our proposed method. Both quantitative and qualitative results on noise-free and noisy benchmark datasets are provided. $Cones$, $Teddy$, $Tsukuba$ and $Venus$, as noisy-free depth maps, are extracted from the Middlebury 2001 and 2003 datasets~\cite{A-taxonomy-and-evaluation-of-dense-two-frame-stereo-correspondence-algorithms:IJCV-2002}\cite{High-accuracy-stereo-depth-maps-using-structured-light:CVPR-2003}. $Art$, $Books$, $Laundry$ and $Reindeer$ as noisy depth maps, are collected from the Middlebury 2005 dataset~\cite{Evaluation-of-cost-functions-for-stereo-matching:CVPR-2007}\cite{Learning-conditional-random-fields-for-stereo:CVPR-2007}. Furthermore, to further evaluate our proposed method, $Jadeplant$, $Motorcycle$, $Playtable$ and $Flower$, as noisy depth maps, are extracted from the Middlebury 2014 dataset~\cite{Middlebury_2014_dataset}. Additionally, we also demonstrate test results of the Laserscan dataset ($Scan 21$, $Scan 30$ and $Scan 42$) provided by Aodha \etal~\cite{Patch-based-synthesis-for-single-depth-image-super-resolution:ECCV-2012}.

\textbf{Baseline Methods:} We compare our methods with the following five categories of methods:
\begin{enumerate}
\item State-of-the-art single DSR methods: Aodha \etal~\cite{Patch-based-synthesis-for-single-depth-image-super-resolution:ECCV-2012}, Hornacek \etal~\cite{Depth-super-resolution-by-rigid-body-self-similarity-in-3d:CVPR-2013}, Ferstl \etal~\cite{Varitional-depth-superresolution-using-example-based-edge-representations:ICCV-2015} and Xie \etal~\cite{edge-guided-single-depth-image-super-resolution:TIP-2016};
\item State-of-the-art color guided DSR methods: Park \etal~\cite{Park:high-quality-depth-map-upsampling-for-3D-TOF-cameras:ICCV-2011}, Yang \etal~\cite{Yang:Depth-Recovery-Using-an-Adaptive-Color-Guided-Auto-regressive-Model:ECCV2012}, Ferstl \etal~\cite{Image-guided-depth-upsampling-using-anisotropic-total-generalized-variation:ICCV-2013}, Kiechle \etal~\cite{Kiechle:joint-intensity-and-depth-co-sparse-analysis-model:ICCV-2013} ($JID$);
\item Single color image super resolution approaches: Zeyde \etal\cite{On-single-image-scale-up-using-sparse-representations:CS-2012}, Yang \etal\cite{Image-super-resolution-via-sparse-representation:TIP-2010} and Timofte \etal~\cite{Anchored-neighborhood-regression-for-fast-example-based-super-resolution:ICCV-2013}, including two kinds of methods: Global Regression ($GR$) and Anchored Neighborhood Regression ($ANR$), the neighborhood embedding methods proposed by Bevilacqua \etal~\cite{Bevilacgua:Low-complexity-Single-image-Super-resolution-based-on-non-negative-neighbor-embedding:BMVC2012}, including $NE+LS$, $NE+NNLS$ and $NE+LLE$, Huang \etal~\cite{Huang:Single-image-super-resolution-from-transformed-self-exemplars:CVPR2015}, Schulter \etal~\cite{schulter:Fast-and-accurate-image-upscaling-with-super-resolution-forests:CVPR2015} and Lai \etal~\cite{lai:deep-laplacian-pyramid-networks-for-fast-and-accurate-super-resolution:cvpr2017};

\item Standard interpolate approaches: Bicubic and Nearest Neighbour ($NN$);
\item State-of-the-art deep convolutional neural networks: Wang \etal~\cite{Wang:Deep-networks-for-image-super-resolution-with-sparse-prior:ICCV2015}, VDSR-Net~\cite{Kim:Accurate-image-super-resolution-using-very-deep-convolutional-networks:CVPR2016}\footnote{For a fair comparison, results of VDSR-Net are obtained using the same number of layers with ours, which means VDSR-Net contains 10 layers for up-sampling factor $\times 2$, contains 20 layers for up-sampling factor $\times 4$, contains 30 layers for up-sampling factor $\times 8$, and contains 40 layers for up-sampling factor $\times 16$}., Song \etal~\cite{song:Learning-depth-super-resolution-using-deep-convolutional-neural-network:ACCV2016}, Hui \etal~\cite{hui:Depth-map-super-resolution-by-deep-multi-scale-guidance:ECCV2016} (MS-Net) and Riegler \etal~\cite{Riegler:ATGV-Net:ECCV2016} (ATGV-Net).
\end{enumerate}

Note that for these deep learning based methods, we retrained the model using depth maps dataset either with the source code provided by the authors or implement those methods by ourself.

\begin{figure*}
\centering
\includegraphics[width=\linewidth]{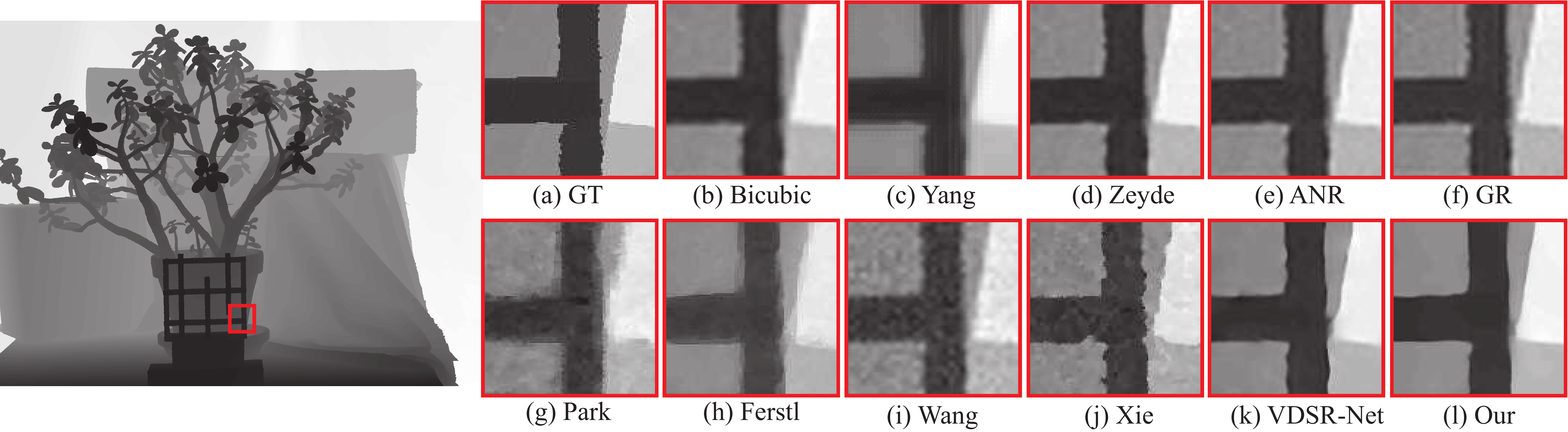} \\
  \caption{Experimental results on the noisy Middlebury 2014 dataset ($Jadeplant$) (up-sampling factor $\times 4$). (a) Ground truth; (b) Bicubic; (c) Yang \etal~\cite{Image-super-resolution-via-sparse-representation:TIP-2010}; (d) Zeyde \etal~\cite{On-single-image-scale-up-using-sparse-representations:CS-2012}; (e) ANR~\cite{Anchored-neighborhood-regression-for-fast-example-based-super-resolution:ICCV-2013}; (f) GR~\cite{Anchored-neighborhood-regression-for-fast-example-based-super-resolution:ICCV-2013}; (g) Park \etal~\cite{Park:high-quality-depth-map-upsampling-for-3D-TOF-cameras:ICCV-2011}; (h) Ferstl \etal~\cite{Image-guided-depth-upsampling-using-anisotropic-total-generalized-variation:ICCV-2013}; (i) Wang \etal~\cite{Wang:Deep-networks-for-image-super-resolution-with-sparse-prior:ICCV2015}; (j) Xie \etal~\cite{edge-guided-single-depth-image-super-resolution:TIP-2016}; (k) VDSR-Net~\cite{Kim:Accurate-image-super-resolution-using-very-deep-convolutional-networks:CVPR2016}; (l) Our results.  \textbf{Best viewed on screen.}}
  \label{fig:jadeplant_noise_x4}
\end{figure*}

\textbf{Error metrics:} In this paper, for quantitative comparison, we use two kinds of error metrics to evaluate the results obtained by our method and other state-of-the-art methods, including: 1) Root Mean Squared Error ($RMSE$); 2). Structure Similarity of Index ($SSIM$). As a frequently used measure, $RMSE$ represents the sample standard deviation of the differences between the obtained HR depth maps and the ground truth. Specifically, $RMSE = \sqrt{\sum_{i=1}^N{(X_{od,i} - X_{gt,i})^2}/N}$, where $X_{od}$ and $X_{gt}$ are the obtained HR depth map and ground truth respectively, $N$ is the number of pixels in the HR depth map. Meanwhile, $SSIM$ is an error metric which evaluates the perceived quality between the obtained HR depth map and the ground truth depth map. We have also computed the percent of errors (percent of pixels that have errors larger than 1 pixel in the disparity map) and the results of percent errors are pretty consistent with the error metrics of $RMSE$ and $SSIM$ reported in our paper.

\subsection{Clean Input}
We evaluate the performance of our method on clean depth maps (there is no noise added in the process of down-sampling). The commonly used $Cones$, $Teddy$, $Tsukuba$ and $Venus$ from the Middlebury stereo dataset are employed. The quantitative results in terms of $RMSE$ and $SSIM$ of up-sampling factors of $\times 2$, $\times 4$ and $\times 8$ are reported in Table~\ref{tab:mb_rmse_2_4_8} and Table~\ref{tab:mb_ssim_2_4_8}. We can clearly see that all the DCNN based methods achieve significant performance leap over other methods under all the up-sampling factors, even though methods~\cite{song:Learning-depth-super-resolution-using-deep-convolutional-neural-network:ACCV2016}\cite{Park:high-quality-depth-map-upsampling-for-3D-TOF-cameras:ICCV-2011}\cite{Image-guided-depth-upsampling-using-anisotropic-total-generalized-variation:ICCV-2013} utilize auxiliary intensity information. Additionally, our proposed method outperforms other DCNN based methods for almost all the up-sampling factors. Note that the VDSR-Net~\cite{Kim:Accurate-image-super-resolution-using-very-deep-convolutional-networks:CVPR2016} utilizes up-sampled (bicubic) depth map as input, and its structure is the same as our DCNN unit. Additionally, for a fair comparison, results of VDSR-Net~\cite{Kim:Accurate-image-super-resolution-using-very-deep-convolutional-networks:CVPR2016} are obtained using the same number of layers with ours, which means VDSR-Net contains 10 layers for up-sampling factor $\times 2$, contains 20 layers for up-sampling factor $\times 4$, and contains 30 layers for up-sampling factor $\times 8$. Obviously, our proposed method outperforms VDSR-Net, which demonstrates the excellent performance of our deeply supervised novel view synthesis and multi-scale fusion strategy. Meanwhile, for ~\cite{lai:deep-laplacian-pyramid-networks-for-fast-and-accurate-super-resolution:cvpr2017}, as shown in Table~\ref{tab:mb_rmse_2_4_8} and Table~\ref{tab:mb_ssim_2_4_8}, we trained two kinds of models, namely $MSLaplas$ (up-sampling factors of $\times 2$, $\times 4$ and $\times 8$ are trained together) and $Laplas$ (up-sampling factors of $\times 2$, $\times 4$ and $\times 8$ are trained separately), we can see that results obtained by our method are better than~\cite{lai:deep-laplacian-pyramid-networks-for-fast-and-accurate-super-resolution:cvpr2017} under up-sampling factors of $\times 2$, $\times 4$ and $\times 8$, which further proves the effectiveness of our method.

As indicated in Table~\ref{tab:mb_rmse_2_4_8} and Table~\ref{tab:mb_ssim_2_4_8}, our deeply supervised novel view synthesis strategy ($Our$) outperforms others for most of the up-sampling factors, and the multi-scale fusion strategy ($Our+MSF$) can handle blocking effect and further improve the results for most up-scale factors, though results of $Our + MSF$ are slightly lower than $Our$ for an up-sampling factor of $\times 2$. Lastly, the depth field statistic prior ($Our+MFS+DFS$) can further improve the results.

Qualitative results are illustrated in Fig.~\ref{result_middlebury} (part of $Cones$ and $Tsukuba$) for an up-sampling factor $\times4$. Obviously, our method produces more visually appealing results. Boundaries generated by our method are sharper and accurate, which show that our method can effectively recover the structure of high-resolution depth maps.

{\footnotesize
\begin{table*}[!h]
\scriptsize
\begin{center}

\caption{Quantitative evaluation. The $SSIM$ is calculated for different SOTA methods on the noisy Middlebury and the Laserscanner datasets under up-sampling factors of $\times{16}$. The best result is highlighted and the second best is underlined.}
\begin{tabular}{ |c|c|c|c|c|c|c|c|c|c|c|c| }
\hline
\multirow{2}{4em}{Method}  & \multicolumn{11}{| c |}{$\times 16$} \\ \cline{2-12}
 &Jadeplant & Motorcycle & Playtable & Flower& Art & Books & Laundry & Reindeer & Scan21 & Scan30 & Scan42\\  \hline

 NN & 0.8812 & 0.9326 & 0.9502 & 0.9437 & 0.8855 & 0.9565 & 0.9362 & 0.9352 & 0.9410 & 0.9485 & 0.9159 \\
 Bicubic & 0.9067 & 0.9577 & 0.9724 & 0.9647 & 0.9258 & 0.9755 & 0.9589 & 0.9614 & 0.9666 & 0.9699 & 0.9446 \\ \hline
 Park et al.~\cite{Park:high-quality-depth-map-upsampling-for-3D-TOF-cameras:ICCV-2011} & 0.9125 & 0.9613 & 0.9782 & 0.9727 & 0.9401 & 0.9792 & 0.9632 & 0.9721 & N/A & N/A & N/A\\
Zeyde et al.~\cite{On-single-image-scale-up-using-sparse-representations:CS-2012} & 0.9067 & 0.9575 & 0.9741 & 0.9651 & 0.9034 & 0.9792 & 0.9632 & 0.9721 & 0.9482 & 0.9644 & 0.9277\\
Yang et al.~\cite{Image-super-resolution-via-sparse-representation:TIP-2010} & 0.9080 & 0.9596 & 0.9750 & 0.9671 & 0.9356 & 0.9804 & 0.9666 & 0.9664 & 0.9675 & 0.9722 & 0.9426 \\
ANR~\cite{Anchored-neighborhood-regression-for-fast-example-based-super-resolution:ICCV-2013} & 0.9056 & 0.9560 & 0.9734 & 0.9636 & 0.8995 & 0.9627 & 0.9421 & 0.9414 &0.9443 & 0.9509 & 0.8970 \\
GR~\cite{Anchored-neighborhood-regression-for-fast-example-based-super-resolution:ICCV-2013} & 0.9056 & 0.9560 & 0.9734 & 0.9636 & 0.9224 & 0.9723 & 0.9564 & 0.9566 & 0.9598 & 0.9644 & 0.9277 \\
JID~\cite{Kiechle:joint-intensity-and-depth-co-sparse-analysis-model:ICCV-2013} & 0.9080 & 0.9641 & 0.9769 & 0.9716 & N/A & N/A & N/A & N/A & N/A & N/A & N/A\\
Wang et al.~\cite{Wang:Deep-networks-for-image-super-resolution-with-sparse-prior:ICCV2015} & 0.9083 & 0.9578 & 0.9708 & 0.9699 & 0.9362 & 0.9742 & 0.9609 & 0.9643 & 0.9746 & 0.9771 & 0.9577\\
VDSR-Net~\cite{Kim:Accurate-image-super-resolution-using-very-deep-convolutional-networks:CVPR2016} & 0.9582 & 0.9639 & 0.9766 & 0.9735 & 0.9372 & 0.9800 & 0.9658 & 0.9683 & 0.9737 & 0.9776 & 0.9547 \\ \hline
$Our$ & 0.9618 & 0.9687 & \underline{0.9810} & \underline{0.9816} & 0.9501 & 0.9847 & 0.9724 & 0.9757 & \underline{0.9817} & 0.9829 & \textbf{0.9757}\\
$Our+MFS$ & \underline{0.9620} & \underline{0.9685} & 0.9809 & 0.9814 & \underline{0.9504} & \underline{0.9848} & \underline{0.9725} & \underline{0.9757} & 0.9815 & \underline{0.9830} & 0.9746 \\
$Our + MFS + DFS$ & \textbf{0.9623} & \textbf{0.9689} & \textbf{0.9811} & \textbf{0.9817} & \textbf{0.9509} & \textbf{0.9850} & \textbf{0.9728} & \textbf{0.9760} & \textbf{0.9819} & \textbf{0.9833} & \underline{0.9752} \\ \hline

\end{tabular}

\label{tab:mb_ls_ssim_16_noise}
\end{center}
\end{table*}
}

\subsection{Noisy Input}\label{sec:noisy_input}
We also evaluate our proposed method on the noisy Middlebury dataset~\cite{Learning-conditional-random-fields-for-stereo:CVPR-2007}\cite{Evaluation-of-cost-functions-for-stereo-matching:CVPR-2007}\cite{Middlebury_2014_dataset} and the Laserscan dataset~\cite{Patch-based-synthesis-for-single-depth-image-super-resolution:ECCV-2012}. To simulate the acquisition process of a Time-of-Flight sensor, depth dependent Gaussian noise is added to the Middlebury dataset ($Art$, $Books$, $Laundry$ $Reindeer$ of the Middlebury 2005 dataset and $Jadeplant$, $Motorcycle$, $Playtable$, $Flower$ of the Middlebury 2014 dataset) and the Laserscan dataset ($Scan 21$, $Scan 30$ and $Scan 42$). Following~\cite{Riegler:ATGV-Net:ECCV2016}, we add depth dependent Gaussian noise to our low-resolution training data $\mathbf{D}^{LR}$ in the form $\theta(d) = \mathcal{N}(0,\delta/d)$, where $\delta = 651$ and $d$ denotes the depth value of each pixel in $\mathbf{D}^{LR}$.

Table~\ref{ls_mb_noisy_x4} and Table~\ref{tab:mb_ls_ssim_4_noise} report the quantitative results in terms of $RMSE$ and $SSIM$ for the up-sampling factor of $\times 4$ with noisy input, from which, we can clearly see that our proposed method outperforms others. We can observe that our method can well eliminate the influence of noise, thus depth maps with smaller $RMSE$ and larger $SSIM$ can be obtained. Note that results of VDSR-Net~\cite{Kim:Accurate-image-super-resolution-using-very-deep-convolutional-networks:CVPR2016} are obtained using the same number of layers with ours for noisy input.

Besides, Fig.~\ref{fig:jadeplant_noise_x4} illustrates the qualitative results of our method ($Jadeplant$) for an up-sampling factor $\times 4$. As shown in Fig.~\ref{fig:jadeplant_noise_x4}, $Jadeplant$ contains complex textures and luxuriant details, which is hard to recover an HR depth map from an LR depth map. Obviously, our method produces more visually appealing results. Boundaries generated by our method are sharper and accurate, which show that our method can effectively recover the structure of high-resolution depth maps.

\begin{figure*}
\centering
\includegraphics[width=\linewidth]{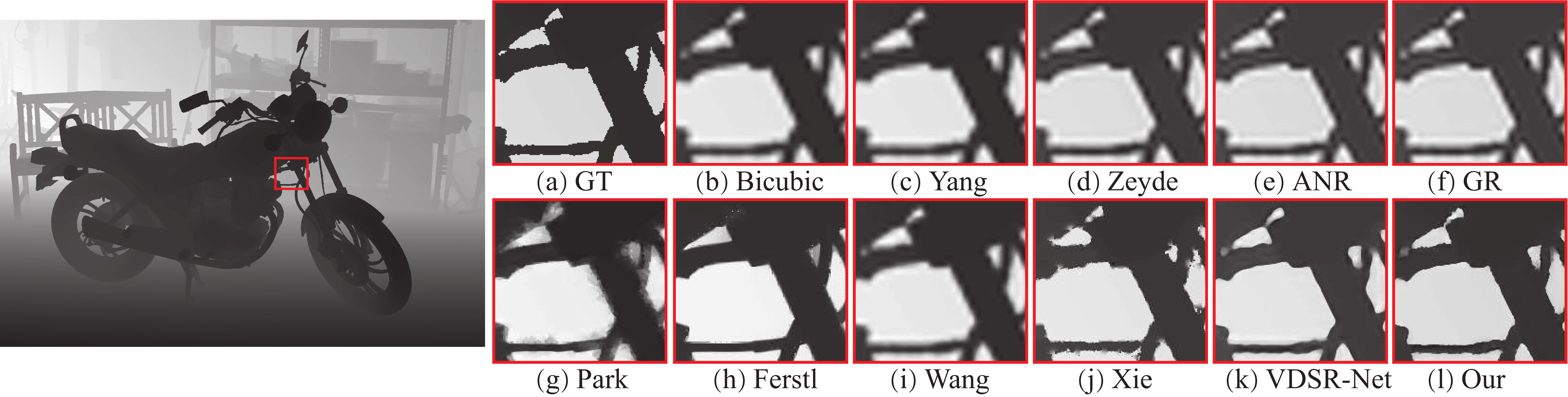} \\
  \caption{Experimental results on the noisy Middlebury 2014 dataset ($Motorcycle$) (up-sampling factor $\times 6$). (a) Ground truth; (b) Bicubic; (c) Yang \etal~\cite{Image-super-resolution-via-sparse-representation:TIP-2010}; (d) Zeyde \etal~\cite{On-single-image-scale-up-using-sparse-representations:CS-2012}; (e) ANR~\cite{Anchored-neighborhood-regression-for-fast-example-based-super-resolution:ICCV-2013}; (f) GR~\cite{Anchored-neighborhood-regression-for-fast-example-based-super-resolution:ICCV-2013}; (g) Park \etal~\cite{Park:high-quality-depth-map-upsampling-for-3D-TOF-cameras:ICCV-2011}; (h) Ferstl \etal~\cite{Image-guided-depth-upsampling-using-anisotropic-total-generalized-variation:ICCV-2013}; (i) Wang \etal~\cite{Wang:Deep-networks-for-image-super-resolution-with-sparse-prior:ICCV2015}; (j) Xie \etal~\cite{edge-guided-single-depth-image-super-resolution:TIP-2016}; (k) VDSR-Net~\cite{Kim:Accurate-image-super-resolution-using-very-deep-convolutional-networks:CVPR2016}; (l) Our results.  \textbf{Best viewed on screen.}}
  \label{fig:motorcycle_x6}
\end{figure*}

\subsection{Large Up-sampling Factor}
Most of the existing work such as~\cite{edge-guided-single-depth-image-super-resolution:TIP-2016}\cite{Varitional-depth-superresolution-using-example-based-edge-representations:ICCV-2015} could only handle depth super-resolution problem under up-sampling factors $\times 2$, $\times 3$ and $\times 4$. In our paper, due to the proposed deeply supervised learning network structure, our method can handle large up-sampling factors effectively. Meanwhile, the multi-scale fusion and depth field statistics can further refine the obtained results. In this section, we demonstrate the qualitative results under large up-sampling factors ($\times 6$, $\times 8$ and $\times 16$). It is a very challenging super-resolution task, \ie, on average inferring $16\times 16 = 256$ depth values from $1$ depth value for $\times 16$.

\begin{figure*}
\centering
\includegraphics[width=\linewidth]{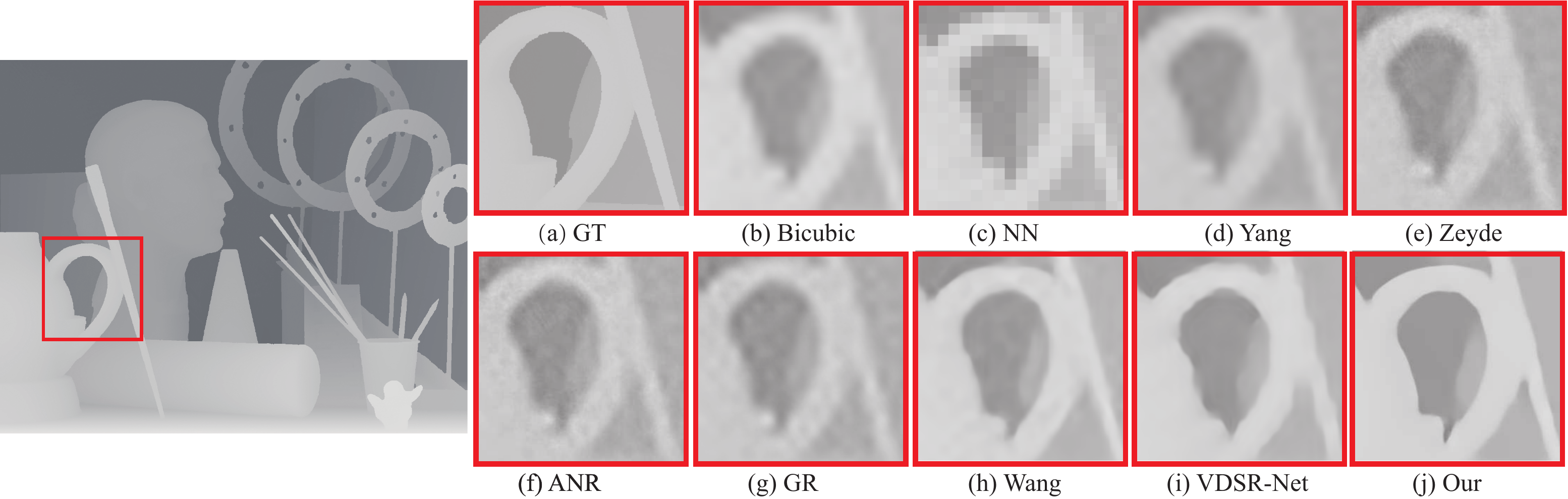} \\
\caption{Experimental results on the noisy Middlebury dataset ($Art$) under an up-sampling factor of $\times 16$. (a) Ground truth; (b) Bicubic; (c) Nearest Neighbor; (d) Yang \etal~\cite{Image-super-resolution-via-sparse-representation:TIP-2010}; (e) Zeyde \etal~\cite{On-single-image-scale-up-using-sparse-representations:CS-2012}; (f) ANR~\cite{Anchored-neighborhood-regression-for-fast-example-based-super-resolution:ICCV-2013}; (g) GR~\cite{Anchored-neighborhood-regression-for-fast-example-based-super-resolution:ICCV-2013}; (h) Wang \etal\cite{Wang:Deep-networks-for-image-super-resolution-with-sparse-prior:ICCV2015}; (i) VDSR-Net~\cite{Kim:Accurate-image-super-resolution-using-very-deep-convolutional-networks:CVPR2016}; (j) Our results.  \textbf{Best viewed on screen.}}
  \label{fig:art_noise_x16}
\end{figure*}

Fig.~\ref{fig:motorcycle_x6} shows the qualitative comparison results of $\times 6$ of $Motorcycle$ from the Middlebury 2014 dataset~\cite{Middlebury_2014_dataset}, which demonstrates the ability of our method in handling general up-sampling factors besides $2^n$ ($\times 2, \times 4, \times, 8, \times 16$). Fig.~\ref{fig:ls_clean_x8} ($Scan 42$ of the Laserscan dataset) shows the qualitative comparison results for an up-sampling factor $\times 8$. Clean low-resolution depth maps are used as input of Fig.~\ref{fig:motorcycle_x6} and Fig.~\ref{fig:ls_clean_x8}. As these figures illustrate, depth maps generated by other competing methods all suffer from heavy edge blur, while our method can obtain results with sharp boundaries and accurate depth edges.

Meanwhile, we also illustrate the qualitative results under large up-sampling factor $\times16$ in Fig.~\ref{fig:art_noise_x16} ($Art$ of the Middlebury dataset). Note that Gaussian noisy described in Section~\ref{sec:noisy_input} is added in the depth input of Fig.~\ref{fig:art_noise_x16}. Obviously, our method can well eliminate the influence of noise and accurately recover the high-resolution depth maps with sharper boundaries, even under a very large up-sampling factor of $\times 16$.

In addition, Table~\ref{ls_mb_noisy_x16} and Table~\ref{tab:mb_ls_ssim_16_noise} demonstrate the quantitative results of our method on $RMSE$ and $SSIM$, respectively. As illustrated in these tables, our method outperforms all competing methods with a margin by achieving smaller $RMSE$ and larger $SSIM$ on all the testing datasets. Note that same depth dependent Gaussian noisy is added to the depth maps as described in Section~\ref{sec:noisy_input}.

\subsection{Ablation Analysis}
As shown in Table~\ref{tab:mb_rmse_2_4_8}, Table~\ref{tab:mb_ssim_2_4_8}, Table~\ref{ls_mb_noisy_x4}, Table~\ref{tab:mb_ls_ssim_4_noise}, Table~\ref{ls_mb_noisy_x16} and Table~\ref{tab:mb_ls_ssim_16_noise}, compared with other state-of-the-art methods, our deeply supervised novel view synthesis strategy ($Our$) can obtain better results. To handle the blocking effect introduced in the resultant depth maps, the multi-scale fusion strategy ($Our$ + $MSF$) is used, and as shown in these tables, the multi-scale fusion strategy ($Our + MFS$) can improve the $RMSE$ and $SSIM$ under most of the testing depth maps. Besides, since all DCNN based methods can only learn the generality of depth super-resolution, we also use depth filed statistic ($DFS$) information (features of each depth map) to further refine the results. As shown in these tables, better $RMSE$ and $SSIM$ results can be obtained for most of the testing depth maps by using depth field statistic ($Our + MFS + DFS$). Therefore, we can conclude that the three components, namely, deeply supervised novel view synthesis strategy, multi-scale fusion strategy and depth field statistics all contribute positively toward the final success of our approach.

\subsection{Influence of the number of layers}
\begin{figure}[h]
  \centering
  \includegraphics[width=\linewidth]{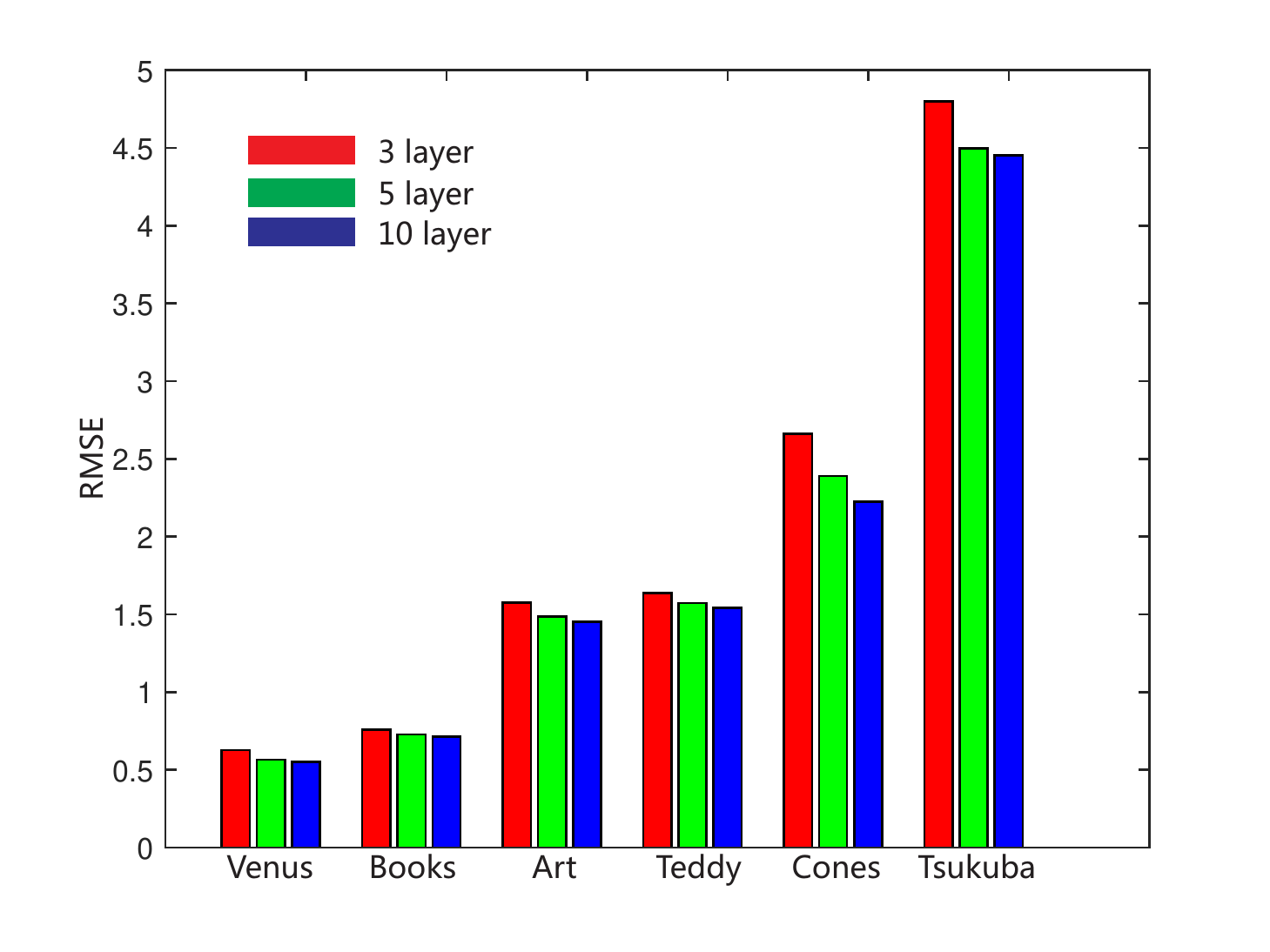} \\
  \caption{$RMSE$ results of our method on the Middlebury dataset (DCNN unit with 3 layers, 5 layers and 10 layers respectively).}
  \label{number_of_layers}
\end{figure}

In this section, we analyze the influence of the number of layers of DCNN unit in novel view synthesis sub-task. Fig.~\ref{number_of_layers} shows the $RMSE$ of our method on the Middlebury dataset, where the number of layers of DCNN unit varies from 3 to 10. We can observe that $RMSE$ decreases with the increase of the number of layers of DCNN unit, however, the difference between 3 layers and 5 layers is larger than the difference between 5 layers and 10 layers. Thus, we can conclude that deeper DCNN generally results in better performance. However, as the net goes deeper, the improvement becomes marginal. Hence, we fix the number of layers of DCNN unit as 10 in our experiments.

\subsection{Running time}
We summarize the computation time for up-sampling different LR depth maps $Art$ to their full resolution ($1376 \times 1104$) in Table~\ref{running_time}. Up-sampling is performed in MATLAB with a TITAN X GPU (Pascal) using MatConvNet~\cite{MatConvNet}. Note that the running time of MS-Net~\cite{hui:Depth-map-super-resolution-by-deep-multi-scale-guidance:ECCV2016} is obtained from their paper which is calculated using a TITAN X GPU.

{\footnotesize
\begin{table}[h]
\scriptsize
\begin{center}

\caption{\upshape Computation time for different methods under different up-sampling factors.}

\begin{tabular}{ |c|c|c|c|c| }
\hline
& $\times 2$ & $\times 4$ & $\times 8$ & $\times 16$   \\  \hline
MS-Net~\cite{hui:Depth-map-super-resolution-by-deep-multi-scale-guidance:ECCV2016} & 0.211 & 0.221 & 0.247 & 0.277 \\  \hline
$Our$ & 0.169 & 0.185 & 0.205 & 0.211 \\  \hline
$Our+MSF$ & 0.315 & 0.331 & 0.351 & 0.357\\ \hline
\end{tabular}

\label{running_time}
\end{center}
\end{table}
}

Besides, the running time of the depth field statistics (DFS) varies a lot depending on different size of input depth maps. For examples, the depth field statistic strategy costs 2.845 seconds on the $Cones$ depth map (370 $\times$ 450 resolution ), while it costs 10.321 seconds on the $Art$ depth map (1376 $\times$ 1104 resolution).

\subsection{Generalization ability}
To evaluate the generalization ability of our network, we use a new training dataset (60 depth maps from the Sintel dataset and 30 depth maps from the synthetic New Tsukuba dataset) to retrain our network, and then test the network on the Middlebury dataset. $RMSE$ results ($\times 8$) are shown in Table\ref{Generalization_Ability} ($Our$\_$Retrain$). $Our$ of Table\ref{Generalization_Ability} shows the results obtained by training data of section~\ref{implementation details}. We can see that $Our$ and $Our\_Retrain$ results are pretty similar to the results reported in Table~\ref{tab:mb_rmse_2_4_8}, which demonstrates the excellent generalization ability of our network.

To further demonstrate our generalization ability on raw depth maps, we show the average $RMSE$ results of the NYU dataset~\cite{NYU_dataset_2}\cite{NYU_dataset_1} in Table~\ref{results_kinect}. The NYU dataset comprises of video sequences of indoor scenes captured by a Kinect. For each set, $5$ depth maps are selected randomly. Note that we use the model trained by training data (115 depth maps from $Middlebury$, $Sintel$ and $Synthetic$ $Tsukuba$ $dataset$) described in our paper to test the NYU dataset. As shown in Table~\ref{results_kinect}, our network model can handle wild and raw depth maps. Note that results shown here are obtained by our deeply supervised novel view synthesis strategy.

\begin{table}[h]
\scriptsize
\begin{center}

\caption{\upshape Experimental evaluation of the generalization ability. ($\times 8$ up-sampling, $RMSE$).}
\begin{tabular}{ |c|c|c|c|c| }
\hline
Method  &  Cones & Teddy & Tsukuba & Venus   \\  \hline
$Our$ & 4.9932 & \textbf{2.8483} & \textbf{9.8526} & 1.1017 \\
$Our$\_$Retrain$ & \textbf{4.9010} & 2.8692 & 10.2326 & \textbf{1.0481} \\
\hline
\end{tabular}

\label{Generalization_Ability}
\end{center}
\end{table}
\vspace{-6mm}

\begin{table}[h]
\scriptsize
\begin{center}

\caption{\upshape Performance evaluation on raw depth maps in the NYU dataset~\cite{NYU_dataset_2}~\cite{NYU_dataset_1} captured with Kinect sensor ($\times 4$, $RMSE$).}
\begin{tabular}{ |c|c|c|c|c| }
\hline
Method & Set$\_$1 & Set$\_$2 & Set$\_$3 & Set$\_$4\\  \hline
Bicubic & 2.7900 & 2.3917  & 2.7752 & 1.8444 \\
VDSR-Net~\cite{Kim:Accurate-image-super-resolution-using-very-deep-convolutional-networks:CVPR2016} & 2.0976 & 1.6226 & 2.0163 & 1.3767 \\
$Our$ & \textbf{2.0399} & \textbf{1.5915} & \textbf{1.9832} & \textbf{1.3513}\\  \hline
\end{tabular}

\label{results_kinect}
\end{center}
\end{table}

\section{Conclusions}
In this paper, we propose to represent depth map super-resolution as a series of novel view synthesis sub-tasks, where each sub-task can be efficiently solved in an end-to-end learning manner. The training stage can be performed in parallel, while neither deconvolution nor pre-processing of input depth map is needed. Furthermore, a deeply supervised learning framework is proposed to handle large up-sampling factors ($\times 8, \times 16$), where strong supervisions are directly applied in different stages. To further exploit the feature maps at different stages, a multi-scale fusion strategy has also been introduced. Our framework can handle DSR efficiently without the need for high-resolution intensity images, and both qualitative and quantitative results demonstrate the outstanding performance of our method compared with the state-of-the-art DSR methods. In the future, we plan to investigate how to exploit the color images under the same framework and how to learn a natural and realistic depth map statistics through GAN (Generative adversarial network).



\ifCLASSOPTIONcompsoc
  \section*{Acknowledgments}
\else
  \section*{Acknowledgment}
\fi

This work is supported by National Natural Science Foundation of China (Nos.61672326, 61671387, 61420106007), National 1000 Young Talents Plan of China, ARC Grants (DE140100180, LP100100588), and China Scholarship Council.


\ifCLASSOPTIONcaptionsoff
  \newpage
\fi

%
\begin{IEEEbiography}[{\includegraphics[width=1in,height=1.25in,clip,keepaspectratio]{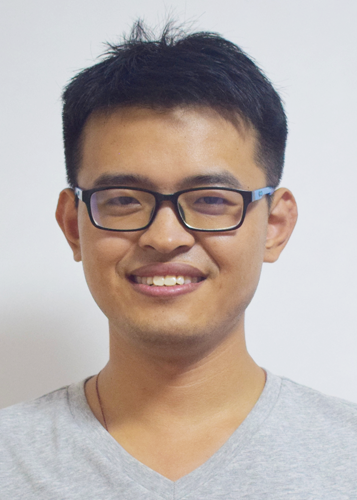}}]
{Xibin Song is currently a senior researcher at Robotics and Autonomous Driving Laboratory (RAL) of Baidu. He obtained his Ph.D. degree in school of Computer Science and Technology from Shandong University, Jinan, China, 2017. He worked as a joint Ph.D. student in the Research School of Engineering at the Australian National University, Canberra, Australia in 2015-2016. He received his B.E. degree in Digital Media and Technology from Shandong University, Jinan, China, in 2011. His research interests include Computer Vision and Augmented Reality.}
\end{IEEEbiography}

\begin{IEEEbiography}[{\includegraphics[width=1in,height=1.25in,clip,keepaspectratio]{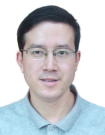}}]
{Yuchao Dai is currently a Professor with School of Electronics and Information at the Northwestern Polytechnical University (NPU). He received the B.E. degree, M.E degree and Ph.D. degree all in signal and information processing from Northwestern Polytechnical University, Xian, China, in 2005, 2008 and 2012, respectively. He was an ARC DECRA Fellow with the Research School of Engineering at the Australian National University, Canberra, Australia from 2014 to 2017 and a Research Fellow with the Research School of Computer Science at the Australian National University, Canberra, Australia from 2012 to 2014. His research interests include structure from motion, multi-view geometry, low-level computer vision, deep learning, compressive sensing and optimization. He won the Best Paper Award in IEEE CVPR 2012, the DSTO Best Fundamental Contribution to Image Processing Paper Prize at DICTA 2014, the Best Algorithm Prize in NRSFM Challenge at CVPR 2017, the Best Student Paper Prize at DICTA 2017 and the Best Deep/Machine Learning Paper Prize at APSIPA ASC 2017.}
\end{IEEEbiography}


\begin{IEEEbiography}[{\includegraphics[width=1in,height=1.25in,clip,keepaspectratio]{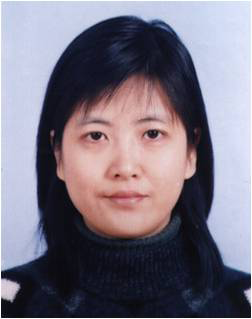}}]
{Xueying Qin is currently a professor of School of Software, Shandong University. She received her Ph.D. from Hiroshima University of Japan in 2001, and M.S. and B.S. from Zhejiang University and Peking University in 1991 and 1998, respectively. Her main research interests are augmented reality, video-based analyzing, rendering, and photo-realistic rendering.}
\end{IEEEbiography}




\end{document}